\documentclass[10pt,twocolumn,letterpaper]{article}

\usepackage{wacv}              % To produce the CAMERA-READY version
\definecolor{wacvblue}{rgb}{0.21,0.49,0.74}
\usepackage[pagebackref,breaklinks,colorlinks,allcolors=wacvblue]{hyperref}

\usepackage[utf8]{inputenc} % allow utf-8 input
\usepackage[T1]{fontenc}    % use 8-bit T1 fonts
\usepackage{hyperref}       % hyperlinks
\usepackage{url}            % simple URL typesetting
\usepackage{booktabs}       % professional-quality tables
\usepackage{amsfonts}       % blackboard math symbols
\usepackage{nicefrac}       % compact symbols for 1/2, etc.
\usepackage{microtype}      % microtypography
\usepackage{xcolor}         % colors

\usepackage{mathtools}
\usepackage{array}
\newcolumntype{C}[1]{>{\centering\arraybackslash}p{#1}}

\usepackage{amsmath}
\usepackage{dsfont}
\usepackage{multirow}
\usepackage{textcomp}

\usepackage{siunitx}
\usepackage{tabularx}
\usepackage{booktabs}
\usepackage{enumitem}

\newcolumntype{Y}{>{\centering\arraybackslash}X}

\def\subcaptionstyle{subcaptionbelow}
\newcommand{\dynamiccaption}[2]{
    \ifx\subcaptionstyle\subcaptionabove
        \setlength{\belowcaptionskip}{10pt}
        #1
    \fi
    #2
    \ifx\subcaptionstyle\subcaptionbelow
        #1
    \fi
}

\usepackage{glossaries}
\newacronym{RS}{RS}{Remote sensing}
\newacronym{CV}{CV}{computer vision}
\newacronym{MLC}{MLC}{multi-label classification}
\newacronym{SLC}{SLC}{single-label classification}
\newacronym{SSL}{SSL}{self-supervised learning}
\newacronym{MoCoV2}{MoCoV2}{Momentum Contrast v2}
\newacronym{k-NN}{k-NN}{k-nearest neighbors}
\newacronym{RRC}{RRC}{RandomResizeCrop}
\newacronym{RR90}{RR90}{RandomRotate90}
\glsdisablehyper

\AtEndPreamble{
    \usepackage[capitalize]{cleveref}
    \crefname{section}{Sec.}{Secs.}
    \Crefname{section}{Section}{Sections}
    \crefname{table}{Tab.}{Tabs.}
    \Crefname{table}{Table}{Tables}
    \crefname{figure}{Fig.}{Figs.}
    \Crefname{figure}{Figure}{Figures}
}

\usepackage[slc=off, subrefformat=parens]{subcaption}
\ifx\subcaptionstyle\subcaptionbelow
\else
\fi

\def\wacvPaperID{2506} % *** Enter the WACV Paper ID here
\def\confName{WACV}
\def\confYear{2026}

\title{Rank-based Geographical Regularization: Revisiting Contrastive Self-Supervised Learning for Multispectral Remote Sensing Imagery}

\author{Tom Burgert\textsuperscript{1,2}, Leonard Hackel\textsuperscript{1,2}, Paolo Rota\textsuperscript{3}, Begüm Demir\textsuperscript{1,2} \\
BIFOLD\textsuperscript{1}, TU Berlin\textsuperscript{2},  University of Trento\textsuperscript{3} \\
{\tt\small \{t.burgert,l.hackel,demir\}@tu-berlin.de}, {\tt\small paolo.rota@unitn.it}
}

\begin{document}
\maketitle
\begin{abstract}
Self-supervised learning (SSL) has become a powerful paradigm for learning from large, unlabeled datasets, particularly in computer vision (CV). However, applying SSL to multispectral remote sensing (RS) images presents unique challenges and opportunities due to the geographical and temporal variability of the data. In this paper, we introduce \textit{GeoRank}, a novel regularization method for contrastive SSL that improves upon prior techniques by directly optimizing spherical distances to embed geographical relationships into the learned feature space. \textit{GeoRank} outperforms or matches prior methods that integrate geographical metadata and consistently improves diverse contrastive SSL algorithms (e.g., BYOL, DINO). Beyond this, we present a systematic investigation of key adaptations of contrastive SSL for multispectral RS images, including the effectiveness of data augmentations, the impact of dataset cardinality and image size on performance, and the task dependency of temporal views. Code is available at https://github.com/tomburgert/georank.

% Our findings challenge prevailing assumptions in multispectral RS contrastive SSL and offer practical insights into designing more effective contrastive SSL pipelines for multispectral RS tasks. 
\end{abstract}    
\section{Introduction}
\label{sec:introduction}

\gls{RS} involves the acquisition of data about the Earth’s surface through sensors on satellites, aircraft, and drones, which capture continuous streams of imagery across various spectral bands. Among the different types of \gls{RS} data, multispectral images from satellite programs such as Landsat \cite{wulder_fifty_2022} and Copernicus Sentinel-2 \cite{gascon_copernicus_2017} have been particularly influential. The open data policy adopted by these satellite missions has facilitated the collection of vast quantities of Earth observation data on a daily basis. This availability of public archives has enabled large-scale applications in which the integration of supervised methods from \gls{CV} has driven breakthroughs in \gls{RS} fields like environmental monitoring \cite{yuan_deep_2020}, agriculture \cite{attri_review_2023}, and urban planning \cite{grekousis_artificial_2019}. \vspace{0.2cm} \\
%\gls{RS} involves the acquisition of data about the Earth’s surface through sensors on satellites, aircraft, and drones, providing continuous streams of optical imagery across various spectral bands. Public space programs such as Landsat \cite{wulder_fifty_2022} and Copernicus (e.g., Sentinel-2) \cite{gascon_copernicus_2017} have been crucial in advancing this technology, facilitating the collection of vast quantities of Earth observation data on a daily basis. The availability of public data archives has enabled large-scale applications in which the integration of supervised methods from \gls{CV} has driven breakthroughs in \gls{RS} fields like environmental monitoring \cite{yuan_deep_2020}, agriculture \cite{attri_review_2023}, and urban planning \cite{grekousis_artificial_2019}. \vspace{0.2cm} \\
\noindent Despite the success of methods inspired from \gls{CV}, it is worth noting that \gls{RS} differs from traditional \gls{CV} domains in several ways \cite{rolf_position_2024}. While varying spatial and spectral resolutions as well as complex acquisition conditions may introduce different challenges that are not typically encountered in \gls{CV}, the availability of zero-cost metadata (e.g., location, time) also enables opportunities in the design of methods in \gls{RS}. Following the recent advances in \gls{CV}, the rise of \gls{SSL} has further expanded the potential of methods in \gls{RS} by leveraging vast amounts of unlabeled data.\vspace{0.2cm} \\
While recent \gls{SSL} approaches in \gls{RS} have introduced domain-specific adaptations such as temporal views for contrastive learning \cite{manas_seasonal_2021}, integrating geographical knowledge \cite{jean_tile2vec_2019}, \cite{ayush_geography-aware_2021}, and masked image modeling for temporal and spectral reconstruction \cite{cong_satmae_2022}, \cite{li_s2mae_2024}, critical aspects remain underexplored. Among approaches that integrate geographical metadata, Tile2Vec \cite{jean_tile2vec_2019} demonstrated that spatial proximity could act as a self-supervision signal, but this predates modern contrastive frameworks. Within contrastive \gls{SSL} methods, existing attempts to integrate geographical information rely on Euclidean distances in a two-stage process \cite{ayush_geography-aware_2021}, which limits their ability to capture Earth’s true spherical structure. To overcome this, we propose \textit{GeoRank}, the first plug-in geographical regularization for contrastive \gls{SSL} in \gls{RS}, which optimizes spherical distances through a rank-based formulation. Unlike previous methods, \textit{GeoRank} introduces geolocation as an inductive bias, constraining the learned representations to reflect the intrinsic geographical structure of the data.
Moreover, prior works have not systematically examined the interplay between data augmentation, dataset size, and input image size, nor have they assessed the task dependency of temporal views. To bridge these gaps, we additionally present a systematic study of contrastive \gls{SSL} adaptations for multispectral \gls{RS} images, establishing new best practices for their application. The main contributions of this work are as follows:

\begin{itemize}[leftmargin=2.4em]
\item  \textbf{Geographical Regularization:} We propose \textit{GeoRank}, the first plug-in geographical regularization for contrastive \gls{SSL} in \gls{RS}, formulated as a rank-based approach which optimizes spherical distances rather than relying on a two-stage process with Euclidean approximations \cite{ayush_geography-aware_2021}. \textit{GeoRank} consistently outperforms or matches prior contrastive methods that integrate geographical metadata and generalizes across multiple contrastive SSL frameworks, with consistent performance gains. %”, improves different contrastive \gls{SSL} algorithms, and integrates seamlessly with RS-specific SSL methods such as temporal and multi-modal contrastive SSL (e.g., CROMA), where it provides additional gains.
% Our approach enhances the alignment of geographically proximate images within the learned feature space, improving downstream generalization when training and evaluation split share similar geographical domains.
\item  \textbf{Data Augmentation:} We demonstrate that the standard augmentation techniques adopted from \gls{CV} contrastive \gls{SSL} are suboptimal for multispectral RS images \cite{jung_contrastive_2022}, \cite{wang_ssl4eo-s12_2023}, \cite{wang_self-supervised_2022}. Through a comprehensive ablation study, we identify a set of augmentation techniques that is better suited to multispectral \gls{RS} images and yield significant performance gains in downstream tasks.
\item  \textbf{Dataset Cardinality:} We challenge the assumption that larger datasets are always necessary for effective contrastive \gls{SSL} on multispectral \gls{RS} images. Contrary to previous findings \cite{manas_seasonal_2021}, \cite{wang_ssl4eo-s12_2023}, our experiments reveal that performance saturation occurs earlier than expected on high-resolution multispectral RS datasets (e.g., Sentinel-2), demonstrating that contrastive \gls{SSL} can be effective even with smaller training sets.
\item  \textbf{Temporal Views:} We provide the first empirical analysis showing that the effectiveness of temporal views in contrastive \gls{SSL} depends on the downstream task. While previous studies \cite{manas_seasonal_2021} assume a general benefit, our findings reveal that temporal views can have varying, and sometimes negative, effects depending on the nature of the task.
\item  \textbf{Image Size:} We challenge the assumption that large image sizes are always necessary for effective contrastive \gls{SSL} on multispectral \gls{RS} images \cite{wang_ssl4eo-s12_2023}. Through controlled experiments, we show that reducing input size during pre-training does not always degrade downstream performance, suggesting that computationally efficient training strategies can be adopted without significant loss in accuracy.
\end{itemize}

\section{Related Work}
\label{sec:related_work}

\textbf{Self-Supervised Learning.} Self-supervised learning (SSL) is a prominent paradigm in visual representation learning that aims to learn generalized representations from unlabeled data through learning signals from within the data itself. The two most prominent approaches include contrastive \gls{SSL} and reconstruction-based \gls{SSL} (i.e., masked image modeling). Contrastive approaches encourage the representations of positive pairs of images (e.g., two augmented views of the same image) to be similar and the representations of negative pairs (views of different images) to be dissimilar. Following the pioneering work of SimCLR \cite{chen_simple_2020}, subsequent approaches like MoCo \cite{he_momentum_2020} have improved negative sample generation through a memory bank of negatives. Later works introduced contrastive-like frameworks that avoid reliance on negative pairs by employing asymmetric architectures, e.g. BYOL \cite{grill_bootstrap_2020}, SimSiam \cite{chen_exploring_2021}, DINO \cite{caron_emerging_2021}. Reconstruction-based approaches involve masking a portion of an image and training a model to predict the masked regions based on the visible context. Popular approaches include MAE \cite{he_masked_2022}, SimMIM \cite{xie_simmim_2022}, and BEiT \cite{bao_beit_2022}. There exist structural differences in the learned representations of contrastive and reconstruction-based approaches \cite{xie_revealing_2023}. Contrastive approaches have more inductive bias and learn representations that are more similar to supervised learning. As a consequence, they perform better in retrieval tasks \cite{caron_emerging_2021}. In contrast, reconstruction-based approaches offer more flexibility, and therefore, scale well to large data archives \cite{he_masked_2022}, \cite{singh_effectiveness_2023}. Nonetheless, they can require significant efforts in fine-tuning to be useful for downstream tasks. \vspace{0.2cm} \\
\noindent \textbf{Self-Supervised Learning in Remote Sensing.} Vast amounts of unlabeled archives of satellite images have inspired the development of RS-specific \gls{SSL} methods. Contrastive SSL has been extended by generating different views based on different timestamps of the same geographical location \cite{manas_seasonal_2021}, \cite{mall_change-aware_2023}, predicting cluster assignment based on geographical metadata \cite{ayush_geography-aware_2021}, or, creating contrastive views based on imagery from different data modalities (e.g., sensors) \cite{scheibenreif_self-supervised_2022}, \cite{feng_cross-modal_2023}. Reconstruction-based approaches include the reconstruction of different scales of resolution \cite{reed_scale-mae_2023}, \cite{noman_rethinking_2024}, the extension of masking strategies to the temporal and spectral dimension \cite{cong_satmae_2022}, \cite{li_s2mae_2024}, or utilizing different modalities of data \cite{fuller_croma_2023}, \cite{hackstein_exploring_2024}. Recent works in \gls{RS} \gls{SSL} combine the objectives of contrastive and reconstruction-based approaches \cite{fuller_croma_2023}, \cite{tang_cross-scale_2023}, \cite{muhtar_cmid_2023}, or are based on diffusion models \cite{khanna_diffusionsat_2024}. \vspace{0.2cm} \\ 
\noindent \textbf{Integrating Geographical Metadata.} To the best of our knowledge, only two works directly enhance contrastive \gls{SSL} by incorporating geographical metadata. Ayush et al. \cite{ayush_geography-aware_2021} extend the contrastive \gls{SSL} objective with an additional loss based on correct geographical cluster assignment, pre-computed via k-means clustering over image geolocations. In addition to being a two-stage procedure, their method relies on Euclidean distance in geographic coordinate space, which does not accurately reflect geodesic (i.e., spherical) distances on Earth. More recently, Bourcier et al. \cite{bourcier_learning_2024} apply contrastive learning between features and encoded metadata, but may similarly fail to capture spherical distances between locations. Other works leveraging geographical metadata supervise contrastive \gls{SSL} with global land cover maps \cite{li_geographical_2022} or generate distinct views from spatially adjacent image patches \cite{jean_tile2vec_2019}, \cite{kang_deep_2020}.
A separate line of work learns contrastive embeddings of image-location pairs for tasks such as elevation and environmental regression, but does not train or evaluate visual encoders in isolation for image-only downstream tasks \cite{vivanco_cepeda_geoclip_2023}, \cite{klemmer_satclip_2025}, \cite{dhakal_range_2025}.
\section{Method}
\label{sec:method}

Our proposed plug-in regularization term integrates geographical metadata into contrastive \gls{SSL}, while remaining agnostic to the choice of contrastive framework. Modern contrastive learning approaches aim to learn high-dimensional feature representations where semantically or transformation-induced similar images (positive pairs) are mapped close to each other, while dissimilar ones are mapped farther apart. Given an unlabeled dataset of multispectral \gls{RS} images of size $N$, we denote each image as $\mathbf{x}_i \in \mathbb{R}^{C \times H \times W}$ with an associated GPS coordinate $g_i = (\text{lon}_i, \text{lat}_i)$ provided in radians. \vspace{0.2cm}\\
\noindent Any suitable contrastive \gls{SSL} framework can be described by the general setup of an encoder network $f_{\theta}$ that maps an image $\mathbf{x}_i$ to a lower-dimensional representation $\mathbf{z}_i = f_{\theta}(\text{aug}(\mathbf{x}_i))$, where $\text{aug}(\cdot)$ denotes a stochastic data augmentation function. For training, we process images in mini-batches $B$ of size $K$. The learning objective varies across different contrastive \gls{SSL} methods. \vspace{0.2cm}\\
\noindent\textbf{Negative-sample-based approaches} (SimCLR \cite{chen_simple_2020}, MoCo \cite{he_momentum_2020}): Contrastive loss functions rely on explicit negative pairs to separate dissimilar instances in representation space. In these cases, we can define a generic contrastive loss:

\begin{align} \label{eq:contrastive_loss} \mathcal{L}_{\text {SSL}}(B)=-\frac{1}{K} \sum_{i=1}^{K} \frac{\exp(\mathbf{z}_i \cdot \mathbf{z}^{\prime}_i / \tau)}{\sum_{k=1}^{M} \exp (\mathbf{z}_i \cdot \mathbf{z}_k / \tau)} \end{align}

\noindent where $\mathbf{z}_i$ and $\mathbf{z}_i^{\prime}$ are positive pairs, and $M$ the total number of negative samples, including both in-batch negatives and, if applicable, negatives from a memory queue. The temperature parameter $\tau$ controls the distribution sharpness.\vspace{0.2cm}\\
\noindent\textbf{Predictive consistency-based approaches} (BYOL \cite{grill_bootstrap_2020}, SimSiam \cite{chen_exploring_2021}, DINO \cite{caron_emerging_2021}): These methods avoid explicit negatives and instead enforce consistency between representations learned from different views. Their loss functions can be formulated as:

\begin{align} \label{eq:consistency_loss} \mathcal{L}_{\text{SSL}}(B) = \sum_{i=1}^{K} d_{\text{sim}}(f_{\theta}(\mathbf{z}_i), f_{\xi}(\mathbf{z}_i^{\prime})) \end{align}

\noindent where $f_{\xi}$ represents a target network with slowly updated parameters, and $d_{\text{sim}}(\cdot, \cdot)$ is a similarity function, such as cosine similarity or mean squared error.

\begin{figure*}[th!]
    \centering
    \includegraphics[width=\linewidth]{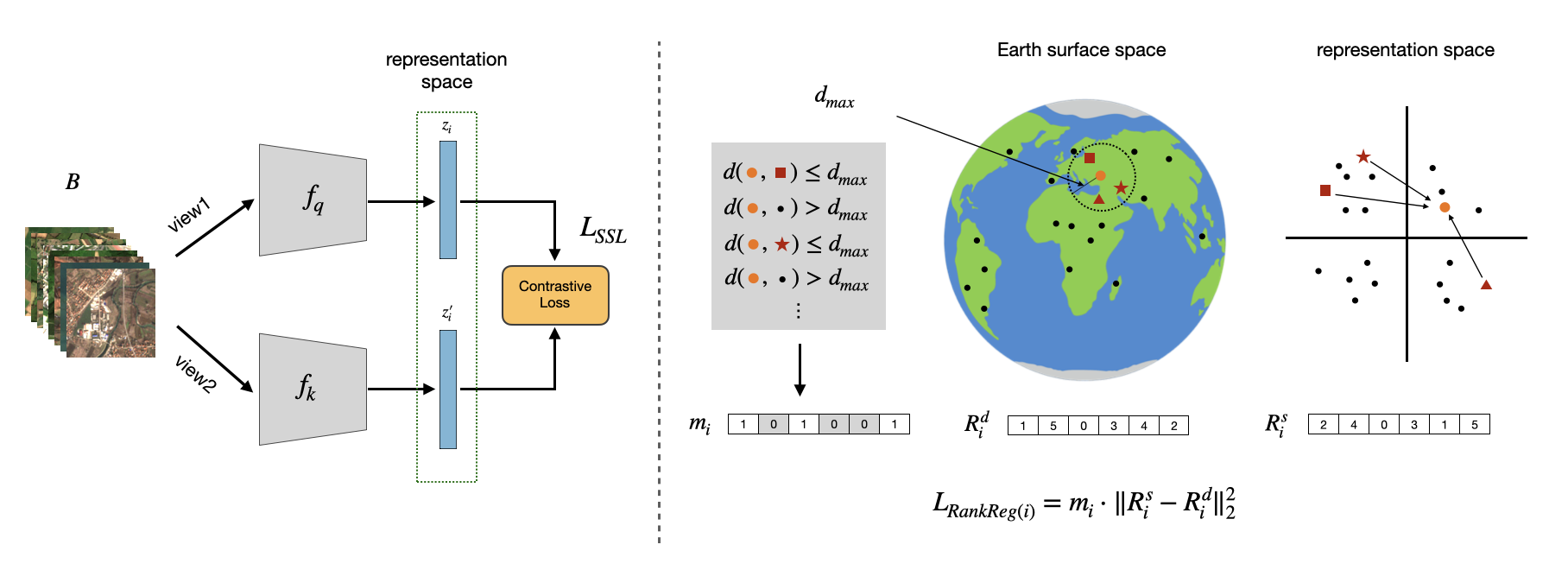}
    \caption{Overview of the proposed plug-in regularization term. \textbf{Left}: A contrastive SSL framework applying contrastive loss $L_{\text{SSL}}$ to the representation space. \textbf{Right}: The proposed regularization loss term $L_{\text{RankReg}}$ for an image $x_i$ in $B$. The order (rank) of distances on the Earth's surface space $\text{R}^\text{d}_i$ (measured by Haversine distance $d$) is used as a label for the order (rank) of distances in representation space $\text{R}^\text{s}_i$. The hyperparameter $m_i$ enables the loss when the distance of two locations on Earth is within the radius $d_{\text{max}}$. For simplicity, we denote the full vector $\mathbf{m}_i$ instead of individual entries $m_{ij}$.}
    \label{fig:method_overview}
    \vspace{-10pt}
\end{figure*}

\subsection{Geographical Regularization}
\label{sec:geoguidance}

Multispectral \gls{RS} images usually come with zero-cost metadata, such as geographic coordinates, which we leverage to improve the representation space. Our method introduces a regularization loss to encourage images from geographically close locations to have similar representations. The motivation behind this regularization is to improve the mid-distance order of the learned representations \cite{muttenthaler_improving_2023}. We define a basic formulation of such a regularization term that incorporates geographical metadata into the representation space as:

\begin{align}
    \label{eq:geomse}
    \mathcal{L}_{\text{Reg}}(B) = \frac{1}{K (K-1)} \sum_{i=1}^K \sum_{\substack{j=1 \\ i \neq j}}^K \| (1 - \mathbf{z}_i \cdot \mathbf{z}_j) - \text{d}(g_i, g_j) \|^2_2
\end{align}

\noindent where $\text{d}(g_i, g_j)$ is the Haversine distance to accurately measure the distance between two locations on a sphere. The full loss of the naive method for geographical regularization is defined as:

\begin{align}
    \label{eq:mse_regularization}
    \mathcal{L}_{\text{GeoBasic}} = \alpha \cdot \mathcal{L}_{\text{SSL}} + (1 - \alpha) \cdot \mathcal{L}_{\text{Reg}}.
\end{align}

\noindent However, direct supervision using raw geographical distances may not align with the learned representation space due to scale and distribution mismatches. Wang and Isola \cite{wang_understanding_2020} highlight uniformity as a key property of the representation space in contrastive learning, but distances in Earth observation data are inherently non-uniform due to land cover heterogeneity of different climate zones and sampling bias due to factors such as high cloud cover. To address this, we propose a rank-based regularization method that preserves relative distance ordering rather than absolute values. By embedding geolocation as a weak supervisory signal, the regularization method introduces a structured inductive bias into contrastive \gls{SSL}, promoting spatial coherence in the representation space that reflects the continuity of Earth’s surface. We calculate the mean squared error (MSE) for the ranks of the distances in representation space and the ranks of the geographical distances. Thus, we define the regularization term RankReg as follows:

\begin{align}
    \label{eq:georank}
    \mathcal{L}_{\text{RankReg}}(B)= \frac{1}{K (K - 1)} \sum_{i=1}^K \sum_{j=1}^{K - 1} m_{ij} \| (\text{R}^\text{s}_i)_j - (\text{R}^\text{d}_i)_j \|^2_2
\end{align}

\noindent where $\text{R}^\text{s}_i = \text{rank}^{-1}(\{ \mathbf{z}_i \cdot \mathbf{z}_k | 1 \leq k \leq K, i \neq k \})$ represents the descending rank order of similarities in the representation space, with lower values assigned to more similar samples and $\text{R}^\text{d}_i = \text{rank}(\{ \text{d}(g_i, g_k) | 1 \leq k \leq K, i \neq k \})$ represents the ascending rank order of geographical distances, with lower values assigned to closer locations. Further, we loosen the geographical constraint by introducing the weighting parameter $m_{ij} = \mathds{1}[\text{d}(g_i, g_j) \leq d_{\text{max}}]$. The weight is only set to 1 if the geographical distance is smaller than a value $d_{\text{max}}$ defining a radius around the location of $g_i$ in which a geographical order of the representation is considered as relevant. Otherwise, the weight is set to 0. When integrated into a contrastive \gls{SSL} framework, we refer to the resulting method as \textit{GeoRank}, with the total loss defined as:

\begin{align}
    \label{eq:rank_regularization}
    \mathcal{L}_{\text{GeoRank}} = \alpha \cdot \mathcal{L}_{\text{SSL}} + (1 - \alpha) \cdot \mathcal{L}_{\text{RankReg}}.
\end{align}
% Thus, the plug-in geographical regularization based on ranks (visualized in \Cref{fig:method_overview}) is defined as:

\section{Experiments}

\label{sec:experiments}
\sisetup{
  table-format=3.0, % Align integer numbers
  detect-weight=true,
  detect-inline-weight=math,
}
\begin{table}
  \centering
  \setlength{\tabcolsep}{4.6pt} % Adjust padding between columns
  \small
  \caption{Overview of the multispectral \gls{RS} datasets used in this work. SSL4EO and BENV2 are used for pre-training, while all datasets except SSL4EO are used for downstream evaluation. Semantic segmentation denoted as SemSeg.}
    
   \begin{tabular}{@{} l c r@{\,}l c c @{}} 
    \toprule
    Dataset & Task & \multicolumn{2}{c}{\#Images} & {Image Size} & {Location} \\
    \midrule
    SSL4EO \cite{wang_ssl4eo-s12_2023} & - & \num{\sim1000} & k & $264\times264$ & Global \\
    BEN-V2 \cite{clasen_reben_2024} & MLC & \num{\sim500} & k & $120\times120$ & Europe \\
    Sen4Agri-ML \cite{sykas_sen4agrinet_2021} & MLC & \num{\sim40} & k & $120\times120$ & Europe \\
    EuroSAT \cite{helber_eurosat_2019} & SLC & \num{\sim27} & k & $64\times64$ & Europe \\
    So2Sat \cite{zhu_so2sat_2020} & SLC & \num{\sim600} & k & $32\times32$ & Global \\
    CashewPlant \cite{lacoste_geo-bench_2023} & SemSeg & \num{\sim2} & k & $256\times256$ & Africa \\
    \bottomrule
  \end{tabular}
  \label{tab:datasets_pretraining_downstream}
\end{table}

\textbf{Datasets.} Throughout our experiments, we use the SSL4EO-S12 dataset \cite{wang_ssl4eo-s12_2023} and BigEarthNet-v2.0 (BEN-V2) \cite{clasen_reben_2024} for pre-training. Downstream performance is evaluated on the \gls{SLC} datasets So2Sat \cite{zhu_so2sat_2020} and EuroSAT (with additional atmospheric correction) \cite{helber_eurosat_2019}, on the \gls{MLC} datasets BEN-V2 and Sen4Agri-ML \cite{sykas_sen4agrinet_2021}, as well as on the semantic segmentation dataset CashewPlant \cite{lacoste_geo-bench_2023} (see \Cref{tab:datasets_pretraining_downstream}). These datasets were selected to span a diverse range of spatial resolutions and geographic extents. In total, up to seven downstream tasks are considered. For So2Sat, we adopt two official splits (So2Sat-rand, So2Sat-block), and for Sen4Agri-ML, we evaluate on both the random (S4A-rand) and tile-based (S4A-tiles) split. Note that Sen4Agri-ML is a multi-label classification dataset derived from the semantic segmentation dataset Sen4AgriNet. All datasets consist of Level-2A (L2A) Sentinel-2 multispectral imagery obtained from the public Copernicus archive. Additional details on dataset specifications and Sentinel-2 characteristics are provided in the supplementary material. \vspace{0.2cm} \\
\textbf{Metrics.} We report performance for the \gls{SLC} datasets and the semantic segmentation dataset in accuracy macro (Acc-mac) and for the \gls{MLC} datasets in mean average precision macro (AP-mac). Avg. Result denotes the average on all six classification downstream tasks. Each score represents the mean of five independent runs with different seeds. \vspace{0.2cm} \\
\begin{figure*}[th!]
    \centering
    \def\segnoisefrac{.32}
    \begin{subfigure}{\segnoisefrac\linewidth}
        {\includegraphics[width=\linewidth]{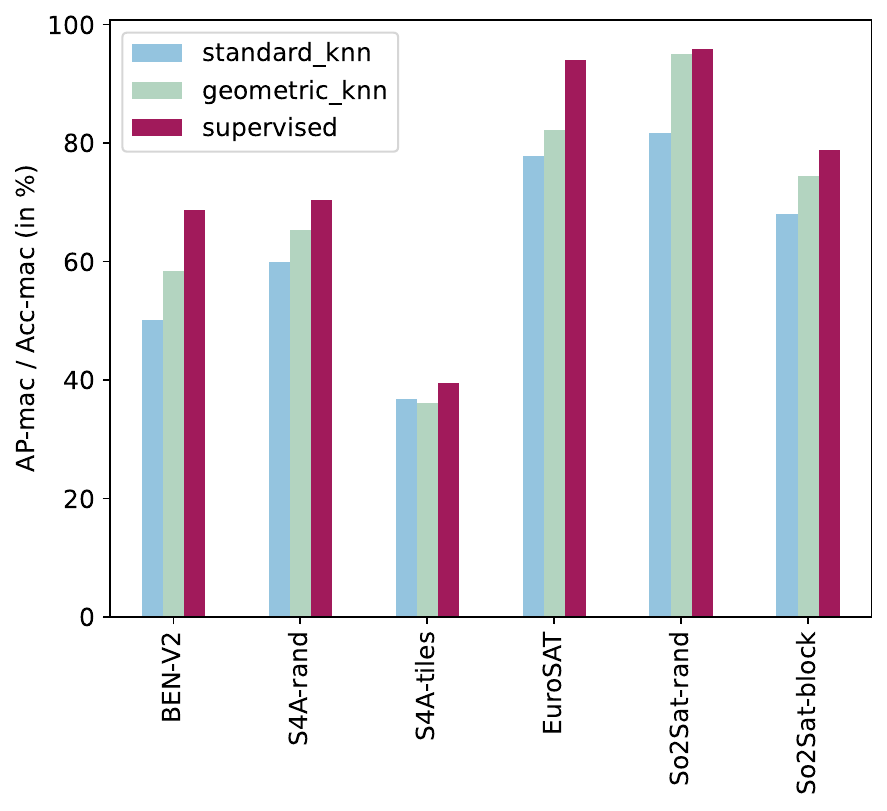}}
        {\caption{}\label{subfigure:ssl4eo_a}}
    \end{subfigure}
    \begin{subfigure}{\segnoisefrac\linewidth}
        {\includegraphics[width=\linewidth]{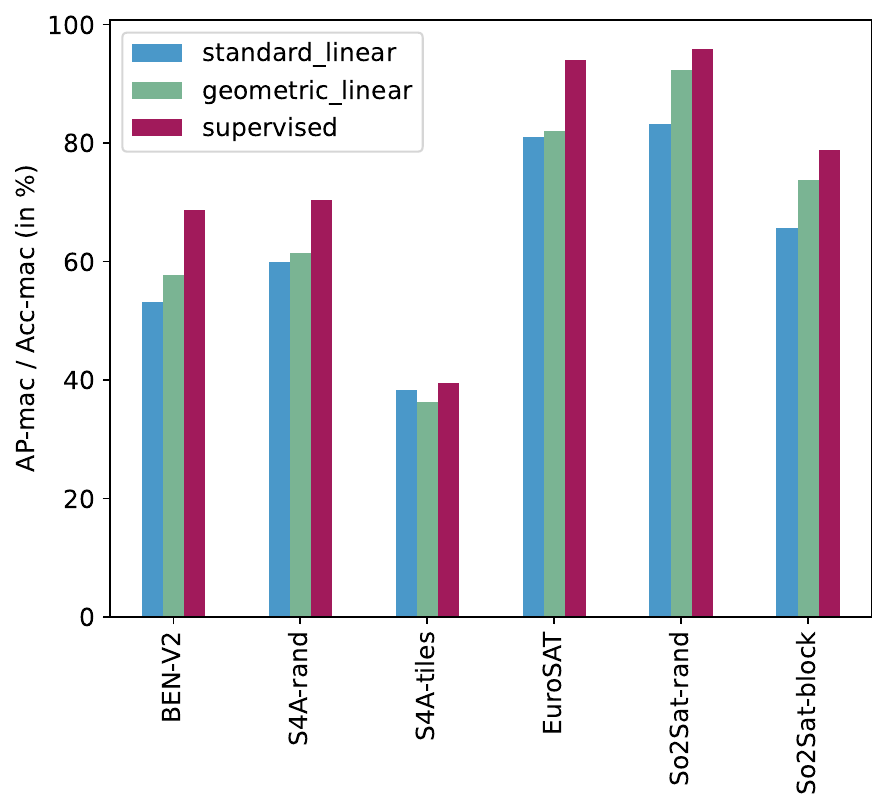}}
        {\caption{}\label{subfigure:ssl4eo_b}}
    \end{subfigure}
    \begin{subfigure}{\segnoisefrac\linewidth}
        {\includegraphics[width=\linewidth]{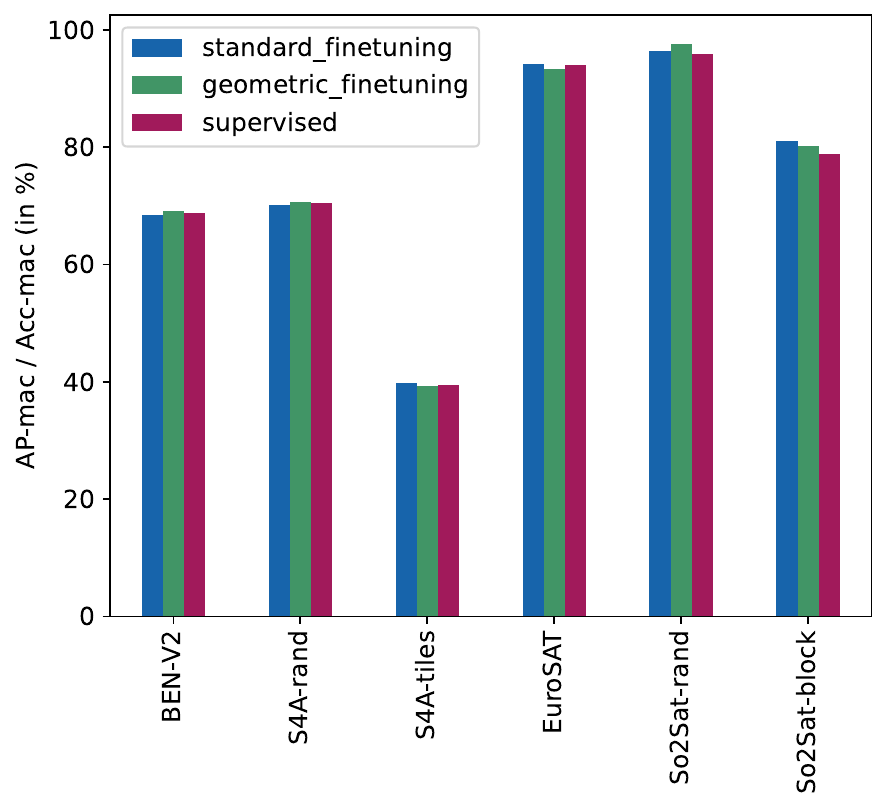}}
        {\caption{}\label{subfigure:ssl4eo_c}}
    \end{subfigure}
    \caption{Performance comparison between the standard augmentation pipeline (blue), the geometric augmentation pipeline (green) and supervised training (red) on six classification downstream tasks when pre-training on SSL4EO: (a) evaluated with the \gls{k-NN} evaluation protocol, (b) evaluated with the linear evaluation protocol, (c) evaluated with the fine-tuning protocol.}
    \label{fig:dataaug_comparison_ssl4eo}
\end{figure*}
\noindent\textbf{Implementation Details.} Given the success of lightweight CNN architectures for \gls{RS} classification image tasks, we base our experiments on a ResNet18 \cite{he_deep_2016}. We evaluate classification downstream performance primarily via \gls{k-NN} classification, alongside linear evaluation and fine-tuning. We evaluate semantic segmentation downstream performance through a UPerNet \cite{xiao_unified_2018}. Unless stated otherwise, the default augmentation pipeline includes \gls{RRC}, horizontal flip, and \gls{RR90}. For SSL4EO, images are center-cropped to $120\times120$ (except when varying pre-training and downstream sizes) to reduce computational cost. The main \textit{GeoRank} experiments and the systematic investigation of key adaptations for contrastive SSL are based on MoCoV2 \cite{chen_improved_2020}. Experiments for extending different \gls{SSL} algorithms use the backbone algorithm specified by the respective method. Since argsort-based rankings are not differentiable, we use a differentiable approximation of the rank function from the FastSoftSort library \cite{blondel_fast_2020}. A comprehensive list of hyperparameters is provided in the supplemental material.

% \begin{table}
%   \centering
%   \setlength{\tabcolsep}{5.7pt} % Adjust padding between columns
%   \small
%   \caption{Overview of the \gls{RS} pretaining (PT) and downstream (DS) datasets used in this work.}
    
%    \begin{tabular}{@{} l c r@{\,}l c c c @{}} 
%     \toprule
%     Dataset & Task & \multicolumn{2}{c}{\#Images} & {Image Size} & {Location} & Use \\
%     \midrule
%     SSL4EO \cite{wang_ssl4eo-s12_2023} & - & \num{\sim1000} & k & $264\times264$ & Global & PT \\
%     BEN-V2 \cite{clasen_reben_2024} & MLC & \num{\sim500} & k & $120\times120$ & Europe & PT/DS \\
%     Sen4Agri-ML \cite{sykas_sen4agrinet_2021} & MLC & \num{\sim40} & k & $120\times120$ & Europe & DS \\
%     EuroSAT \cite{helber_eurosat_2019} & SLC & \num{\sim27} & k & $64\times64$ & Europe & DS \\
%     So2Sat \cite{zhu_so2sat_2020} & SLC & \num{\sim600} & k & $32\times32$ & Global & DS \\
%     \bottomrule
%   \end{tabular}
%   \label{tab:datasets_pretraining_downstream}
% \end{table}

\subsection{Data Augmentation}
\label{sec:dataaug}

Common to many contrastive \gls{SSL} works on multispectral \gls{RS} images is the adoption of the standard set of hyperparameters developed for \gls{CV}, above all the data augmentation pipeline \cite{jung_contrastive_2022}, \cite{wang_self-supervised_2022}, \cite{wang_ssl4eo-s12_2023}. We argue that the specific properties of multispectral \gls{RS} images need to be reflected in the selection of an adequate set of data augmentation techniques for contrastive \gls{SSL} tasks. Previous works overlooked this deficiency by evaluating models solely via fine-tuning. Therefore, we evaluate the performance on six different classification downstream tasks with two different augmentation pipelines: standard and geometric. The standard pipeline consists of \gls{RRC}, ColorJitter (only applying Contrast and Brightness adjustments as Hue and Saturation are not defined for more than 3 channels), GaussianBlur, GrayScale and horizontal flips. The geometric pipeline is composed of \gls{RRC}, Flip (horizontally as well as vertically) and \gls{RR90}. 
\begin{table}[]
  \centering
  \small
  \caption{Ablation study of adding augmentation techniques with probability \num{0.2} and parameter strength $\beta$ to the geometric data augmentation baseline. Each score reflects the average improvement in the \gls{k-NN} protocol over six classification downstream tasks in comparison to the geometric baseline when pre-training on SSL4EO. Base Val. refers to the fixed hyperparameter prior to scaling by $\beta$.}

  \setlength{\tabcolsep}{9pt} % Adjust padding between columns
  \begin{tabularx}{\columnwidth}{@{} l *{3}{S[table-format=1.2]} r @{}}
    \toprule
    Augmentation & {$\beta$} & {$2 \beta$} & {$3 \beta$} & Base Param \\
    \midrule
    Brightness & -0.24 & -0.46 & -0.38 & limit=$0.1$ \\
    Contrast & \bfseries 0.14 & \bfseries 0.14 & -0.03 & limit=$0.1$ \\
    Sharpness & \bfseries 0.17  & \bfseries 0.15 & -0.07 & alpha=$0.1$ \\
    GaussianBlur & -0.07  & 0.06 & -0.08 & sigma=$1.5$ \\
    GaussianNoise & -0.22 & -0.19 & -0.13 & var=$30$ \\
    Solarize & \bfseries 0.27 & {} & {} & threshold=$128$ \\
    Posterize & 0.00 & {} & {} & num-bits=$4$ \\
    Grayscale & -3.78 & {} & {} & {} \\
    \midrule
    RRC & 0.00  & -0.44 & -1.21 & min-scale=$0.2$ \\
    CutOut & 0.01 & 0.07  & \bfseries 0.10 & max-edge=$0.2$ \\
    GridShuffle & \bfseries 0.21 & \bfseries 0.10  & -0.01 & grid-edge=$2$ \\
    Shear & 0.03 & \bfseries 0.12 & 0.01 & angle=$10$ \\
    Translate & 0.02 & \bfseries 0.16 & \bfseries 0.15 & percent=$10$ \\
    \bottomrule
  \end{tabularx}
  \label{tab:augmentation_ablation}
\end{table}

\begin{table*}[t!]
  \centering
  \small
  \caption{Downstream performance (in \%) of different contrastive SSL methods that integrate geographical metadata when pre-training on BEN-V2 evaluated by the \gls{k-NN} protocol.}
  \setlength{\tabcolsep}{5.2pt}
  \begin{tabular}{@{} l c c c c c c c @{}}
    \toprule
    Method & {BEN-V2} & {EuroSAT} & {S4A-rand} & {S4A-tiles} & {So2Sat-rand} & {So2Sat-block} & {CashewPlant} \\
    \midrule
    Baseline \cite{chen_improved_2020} & 58.14 & 82.42 & 65.18 & 35.42 & 93.54 & 72.32 & 31.81 \\
    Tile2Vec \cite{jean_tile2vec_2019} & 54.95 & 73.39 & 63.10 & 35.20 & 81.67 & 66.14 & 34.40 \\
    Ayush et al. \cite{ayush_geography-aware_2021} & 58.41 & 84.34 & \textbf{65.99} & 34.99 & \textbf{94.97} & 73.36 & 34.02 \\
    \midrule
    GeoBasic (ours) & 57.86 & 82.05 & 65.22 & 35.26 & 93.01 & 71.50 & 32.26 \\
    GeoRank (ours) & \textbf{59.19} & \textbf{85.09} & \textbf{65.91} & 35.17 & \textbf{94.95} & \textbf{73.46} & \textbf{34.94} \\
    \midrule
    SatMAE \cite{cong_satmae_2022} & 54.70 & 83.92 & 63.22 & \textbf{37.67} & 87.92 & 68.90 & 19.81 \\
    ScaleMAE \cite{reed_scale-mae_2023} & 43.92 & 64.22 & 53.84 & 35.30 & 54.25 & 47.90 & 27.42 \\
    CrossScaleMAE \cite{tang_cross-scale_2023} & 55.26 & 84.01 & 65.50 & 36.40 & 93.42 & 71.73 & 27.65 \\
    \bottomrule
  \end{tabular}
  \label{tab:geography_aware_results}
\end{table*}
\noindent The results on six classification downstream tasks show that a simple geometric pipeline outperforms the standard pipeline by values of up to \SI{15}{\percent} in the \gls{k-NN} protocol (see \Cref{subfigure:ssl4eo_a}). It is noteworthy that this effect diminishes when more hyperparameters are involved in the evaluation protocol: while the average improvement for linear evaluation is between \SI{3}{\percent} and \SI{5}{\percent} (see \Cref{subfigure:ssl4eo_b}), the differences become marginal when observing the results for the fine-tuning protocol (see \Cref{subfigure:ssl4eo_c}). This pattern highlights that the choice of the augmentation pipeline has the greatest impact when the evaluation protocol directly reflects the structure of the learned representation. As also emphasized by Corley et al. \cite{corley_revisiting_2024}, the \gls{k-NN} protocol is particularly suited for this purpose, as it avoids confounding effects from additional optimization or task-specific tuning. A likely explanation for the observed differences lies in the domain gap between multispectral \gls{RS} images and \gls{CV}: while color-suppressing augmentations such as GrayScale promote shape-biased representations in \gls{CV}, they disrupt critical spectral information in multispectral \gls{RS} images, which is essential for distinguishing semantically similar classes (e.g., different vegetation types). These findings underscore the importance of tailoring data augmentations to the characteristics of multispectral \gls{RS} data and evaluating contrastive \gls{SSL} representations with protocols that do not involve additional hyperparameter tuning. \vspace{0.2cm} \\
\noindent\textbf{Ablation.} We further investigate the effects of individual augmentation techniques within the data augmentation pipeline to derive general recommendations for selecting them when training a contrastive \gls{SSL} algorithm on multispectral \gls{RS} images. We fix the baseline as the geometric pipeline from the initial data augmentation experiments and individually add one data augmentation technique with a probability of \num{0.2} to the augmentation pipeline. For each technique, we conduct experiments with different magnitudes of strength and report the difference in Avg. Result in comparison to the geometric baseline. The results in \Cref{tab:augmentation_ablation} indicate that all channel augmentation techniques except light Contrast or Sharpness adjustments decrease downstream performance. The results particularly emphasize that the application of the \gls{CV}-default augmentation techniques Brightness and Grayscale have a negative impact when used in the augmentation pipeline. On the other hand, a strong application of the geometric augmentation techniques CutOut and Translate as well as a light application of GridShuffle leads to small increases in downstream performance.

\subsection{Geographical Regularization}
\label{sec:ggr}

\begin{figure}[t!]
    \centering
    \def\segnoisefrac{.49}
    \begin{subfigure}{\segnoisefrac\linewidth}
        {\includegraphics[width=\linewidth]{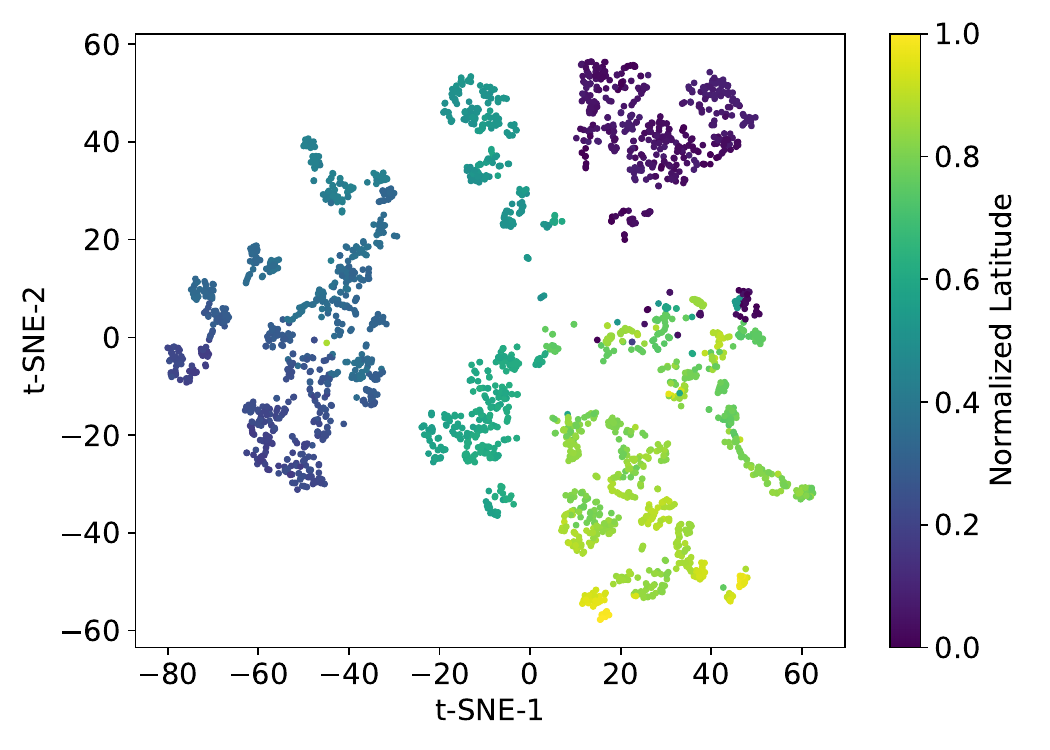}}
    {\caption{}\label{subfigure:georank_latitude}}
    \end{subfigure}
    \begin{subfigure}{\segnoisefrac\linewidth}
        {\includegraphics[width=\linewidth]{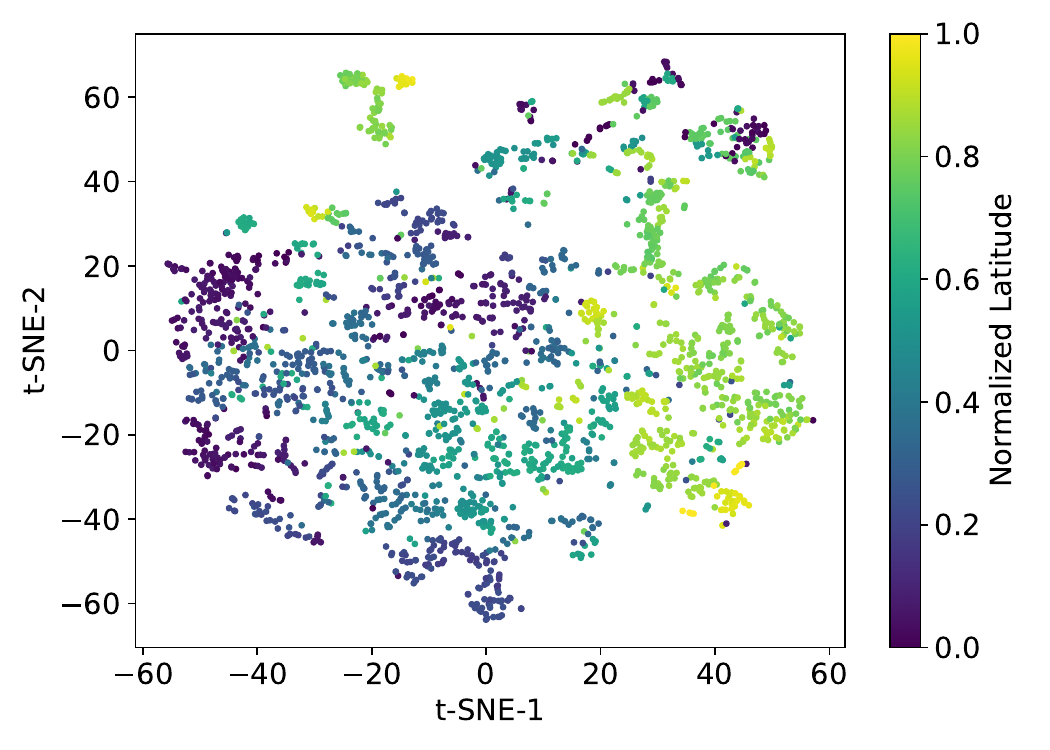}}
        {\caption{}\label{subfigure:baseline_latitude}}
    \end{subfigure}
    \caption{t-SNE of penultimate layer representations for 2560 samples of BEN-V2 after PCA (50 components). Points are colored by normalized latitude in (a) \textit{GeoRank} and (b) Baseline (MoCoV2).}
    \label{fig:t-SNE_analysis_latitude}
\end{figure}

\textbf{Comparison with Existing Work.} \textit{GeoRank} enhances contrastive SSL as a plug-in regularization term that incorporates spatial relationships between image locations through ranking. Unlike clustering-based or absolute-distance approaches (e.g., GeoBasic), \textit{GeoRank} preserves the relative ordering of geographical distances without enforcing rigid alignment in representation space. Tile2Vec \cite{jean_tile2vec_2019} is an early attempt to exploit spatial proximity through contrastive learning. To date, Ayush et al. \cite{ayush_geography-aware_2021} remains the only work that integrates geographical metadata into a modern contrastive framework (MoCoV2). For comparability, we adopt the same backbone in our evaluation. Bourcier et al. \cite{bourcier_learning_2024} also propose to integrate metadata via contrastive learning, but as an implementation of this approach is not available to date, we could not include a quantitative comparison. As summarized in Table~\ref{tab:geography_aware_results}, adding \textit{GeoRank} as a plug-in regularization term consistently improves over both Tile2Vec and the MoCoV2 baseline, and achieves on-par or superior performance compared to Ayush et al., particularly on BEN-V2, EuroSATV2, and CashewPlant. Qualitative analyses based on t-SNE visualizations of BEN-V2 representations from the penultimate layer show that, compared to the MoCo~V2 baseline, \textit{GeoRank} produces smoother spatial organization that reflects relative geographical ordering rather than forming rigid clusters. This qualitative pattern reflects the intended rank-based objective, which preserves these ordering relations without enforcing strict alignment (\Cref{fig:t-SNE_analysis_latitude}, \Cref{fig:t-SNE_analysis_full}). Beyond geography-aware contrastive \gls{SSL} methods, \textit{GeoRank} also achieves stronger performance than recent \gls{RS}-specific masked autoencoder approaches such as SatMAE \cite{cong_satmae_2022}, ScaleMAE \cite{reed_scale-mae_2023} and CrossScaleMAE \cite{tang_cross-scale_2023} across nearly all tasks. While these belong to a different \gls{SSL} family with distinct architectures and objectives and do not integrate geographical metadata, we include them to situate \textit{GeoRank} within the broader landscape of \gls{RS} \gls{SSL}. \vspace{0.2cm} \\
The only case in which \textit{GeoRank} does not outperform prior approaches is S4A-tiles, a dataset characterized by a strong geographical domain shift: the training images originate from Spain, while test images come from France (see supplemental material for details). In such cases, where the training and evaluation sets of a downstream dataset have no geographical overlap, geographic regularization does not provide a clear benefit (also not for Ayush et al. \cite{ayush_geography-aware_2021}). This is intuitive as geographically guided representations improve the sorting of semantically similar locations, but this sorting becomes less relevant when training and test samples are geographically disjoint. Interestingly, in this setting, masked autoencoders exhibit stronger performance, likely due to their flexible feature space that is less constrained by spatial priors. \vspace{0.2cm} \\
% \begin{table}[th!]
%   \centering
%   \small
%     \caption{Downstream performance (in \%) for different contrastive \gls{SSL} algorithms with and without \textit{GeoRank}.}
%     \setlength{\tabcolsep}{5.2pt}
%     \begin{tabular}{@{} l c c c c @{}}
%     \toprule
%     \multirow{2}{*}{\shortstack{Contrastive SSL\\ Algorithm}} & \multirow{2}{*}{\shortstack{BEN-V2\\ \textit{Baseline}}} & \multirow{2}{*}{\shortstack{BEN-V2\\\textit{GeoRank}}} & \multirow{2}{*}{\shortstack{EuroSAT\\\textit{Baseline}}} & \multirow{2}{*}{\shortstack{EuroSAT\\ \textit{GeoRank}}} \\
%     & & \\
%     \midrule
%     SimCLR \cite{chen_simple_2020} & 47.93 & \textbf{54.95} & 65.91 & \textbf{75.06} \\ 
%     BYOL \cite{grill_bootstrap_2020} & 50.40 & \textbf{57.08} & 67.62 & \textbf{79.66} \\ 
%     SimSiam \cite{chen_exploring_2021} & 43.53 & \textbf{57.53} & 52.14 & \textbf{80.10} \\ 
%     DINO \cite{caron_emerging_2021} & 56.61 & \textbf{58.22} & 81.49 & \textbf{81.87} \\ 
%     \bottomrule
%   \end{tabular}
%   \label{tab:different_ssl_backbones}
% \end{table}
\noindent \textbf{Extending Different Contrastive \gls{SSL} Algorithms.} To evaluate the generality of \textit{GeoRank}, we apply it as a plug-in regularization term to four alternative contrastive \gls{SSL} algorithms: SimCLR \cite{chen_simple_2020}, BYOL \cite{grill_bootstrap_2020}, SimSiam \cite{chen_exploring_2021}, and DINO \cite{caron_emerging_2021}. \Cref{tab:different_ssl_backbones} reports downstream classification performance on BEN-V2, EuroSAT, S4A-rand and So2Sat-rand for each algorithm, with and without \textit{GeoRank}. Across all contrastive \gls{SSL} algorithms, the integration of \textit{GeoRank} consistently improves performance, indicating that the plug-in regularization term is agnostic to the underlying contrastive \gls{SSL} algorithm. Notably, even for weaker baselines such as SimSiam, \textit{GeoRank} yields substantial improvements, highlighting its ability to enhance spatial alignment in the learned representations. While stronger methods such as DINO already achieve high baseline performance, the addition of \textit{GeoRank} still provides measurable gains across all datasets, suggesting its utility even in strong-performing contrastive \gls{SSL} algorithm. Additional experiments on the integration of \textit{GeoRank} with other contrastive \gls{SSL} algorithms, including RS-specific methods are reported in the supplemental material.

% Beyond off-the-shelf contrastive \gls{SSL} algorithms, we also tested whether \textit{GeoRank} integrates seamlessly into RS-specific contrastive approaches such as SeCo \cite{manas_seasonal_2021} and CROMA \cite{fuller_croma_2023}. Results, reported in supplemental material (\Cref{tab:extending_rs_ssl_methods}), show that \textit{GeoRank} is fully compatible, with small gains in most cases, suggesting that \textit{GeoRank} can be combined with domain-specific extensions without interfering with their objectives.

\begin{table*}[h!]
  \centering
  \small
    \caption{Downstream performance (in \%) for different contrastive \gls{SSL} algorithms with and without \textit{GeoRank}.}
    
    \setlength{\tabcolsep}{5.2pt}
    \begin{tabular}{@{} l c c c c c c c c @{}}
    \toprule
    \multirow{2}{*}{\shortstack{Contrastive SSL\\ Algorithm}} & \multirow{2}{*}{\shortstack{BEN-V2\\ \textit{Baseline}}} & \multirow{2}{*}{\shortstack{BEN-V2\\\textit{GeoRank}}} & \multirow{2}{*}{\shortstack{EuroSAT\\\textit{Baseline}}} & \multirow{2}{*}{\shortstack{EuroSAT\\ \textit{GeoRank}}} & \multirow{2}{*}{\shortstack{S4A-rand\\\textit{Baseline}}} & \multirow{2}{*}{\shortstack{S4A-rand\\ \textit{GeoRank}}} & \multirow{2}{*}{\shortstack{So2Sat-rand\\\textit{Baseline}}} & \multirow{2}{*}{\shortstack{So2Sat-rand\\ \textit{GeoRank}}} \\
    & & & & \\
    \midrule
    SimCLR \cite{chen_simple_2020} & 47.93 & \textbf{54.95} & 65.91 & \textbf{75.06} & 61.26 & \textbf{63.02} & 83.13 & \textbf{87.40} \\ 
    BYOL \cite{grill_bootstrap_2020} & 50.40 & \textbf{57.08} & 67.62 & \textbf{79.66} & 59.51 & \textbf{64.75} & 77.59 & \textbf{81.09} \\ 
    SimSiam \cite{chen_exploring_2021} & 43.53 & \textbf{57.53} & 52.14 & \textbf{80.10} & 54.42 & \textbf{64.51} & 64.20 & \textbf{91.35} \\ 
    DINO \cite{caron_emerging_2021} & 56.61 & \textbf{58.22} & 81.49 & \textbf{81.87} & 63.72 & \textbf{65.67} & 90.54 & \textbf{93.63} \\ 
    \bottomrule
  \end{tabular}
  \label{tab:different_ssl_backbones}
\end{table*}

% \noindent \textbf{Extending RS-Specific Contrastive SSL Methods.}
% Beyond off-the-shelf contrastive algorithms, we also assess \textit{GeoRank} in the context of \gls{RS}-specific contrastive \gls{SSL} methods that incorporate temporal or multi-modal information. In particular, we add \textit{GeoRank} to Seasonal Contrast (SeCo) \cite{manas_seasonal_2021} and CROMA \cite{fuller_croma_2023}, which represent temporal contrastive learning and multi-modal contrastive learning (i.e., Sentinel-2 with Sentinel-1) respectively. Results across benchmarks show modest but consistent improvements, indicating that \textit{GeoRank} complements these specialized approaches without interfering with their objectives. Consistent with the main experiments, the only case without improvement is the geographical domain shift in S4A-tiles. This highlights that \textit{GeoRank} not only functions as a backbone-agnostic regularizer but also extends established \gls{RS}-specific \gls{SSL} methods, broadening its applicability in the \gls{RS} domain.

\subsection{Dataset Cardinality}
\label{sec:pretraining_size}

\begin{figure}[]
    \centering
    \def\segnoisefrac{.46}
    \begin{subfigure}{\segnoisefrac\linewidth}
        {\includegraphics[width=\linewidth]{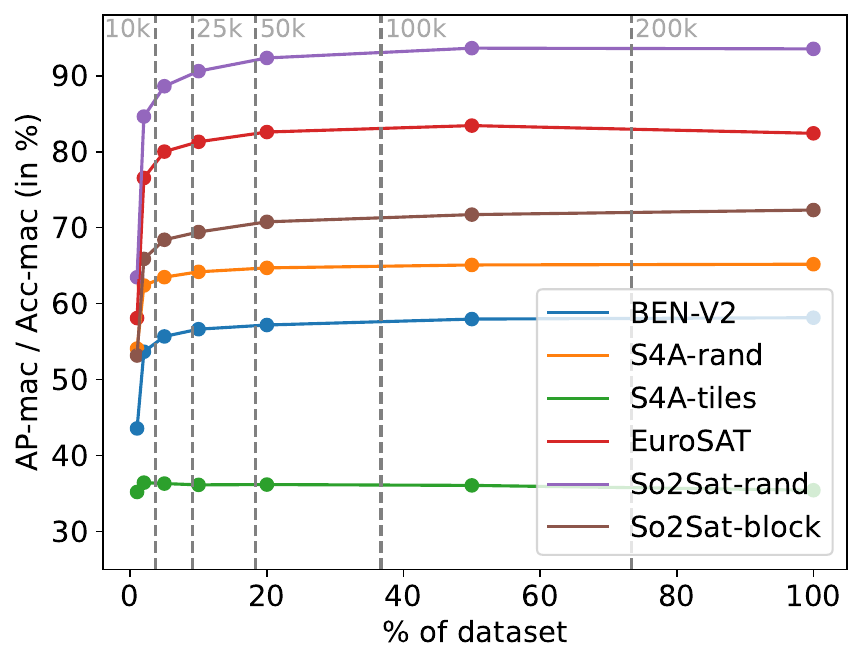}}
        {\caption{}\label{subfigure:cardinality_benv2}}
    \end{subfigure}
    \begin{subfigure}{\segnoisefrac\linewidth}
        {\includegraphics[width=\linewidth]{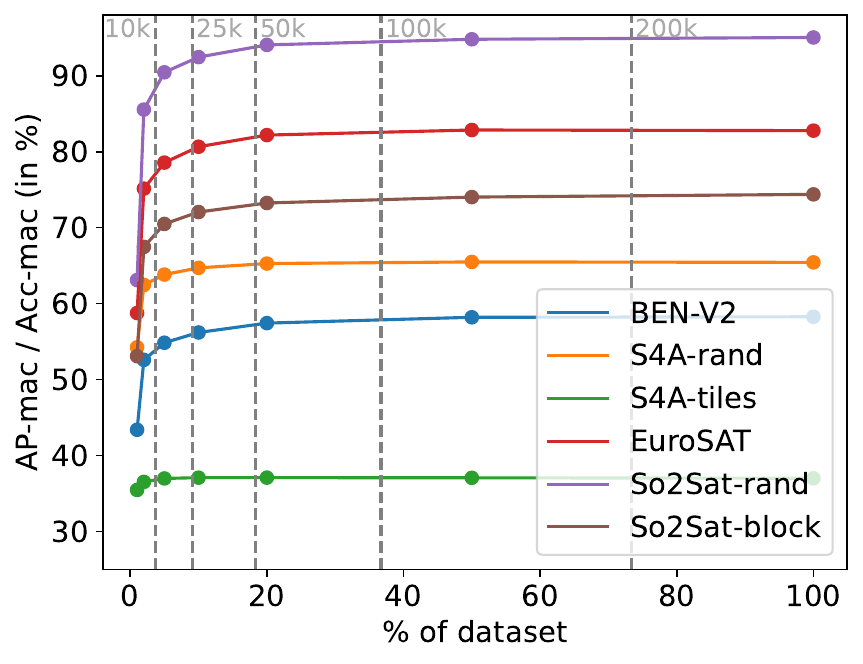}}
        {\caption{}\label{subfigure:cardinality_ssl4eo}}
    \end{subfigure}
    \caption{Performance of different subset sizes of the pre-training dataset (a) BEN-V2 and (b) SSL4EO, evaluated on six classification downstream tasks by \gls{k-NN}.}
    \label{fig:pretraining_size}
\end{figure}

Unlike natural images, the diversity of multispectral \gls{RS} imagery is bounded by the Earth’s surface and satellite spatial resolution (square meters on the ground per pixel). This raises the question of how large a pre-training dataset must be before downstream performance saturates. Prior studies on high-resolution multispectral satellite data (e.g., Sentinel-2) have limitations: Manas et al. \cite{manas_seasonal_2021} tested only two dataset sizes (\num{100000} and \num{1000000}), preventing precise identification of the saturation point, while Wang et al. \cite{wang_ssl4eo-s12_2023} used inconsistent data levels (L1C for pre-training, L2A for downstream) and subsampled the downstream set by \SI{10}{\percent}. Both studies also evaluated only a single downstream task. To address these issues, we ablate the pre-training size for both pre-training datasets and evaluate on six classification diverse downstream tasks, using only L2A-level data throughout. We observe that performance saturates between \num{100000} and \num{200000} pre-training images (\Cref{subfigure:cardinality_benv2}, \Cref{subfigure:cardinality_ssl4eo}); beyond this point, larger datasets yield no additional performance gains. Moreover, in contrast to Wang et al., we find no significant saturation differences across model sizes (\Cref{fig:pretraining_size_model_size}, in the supplemental material).

\subsection{Temporal Views}
\label{sec:seasonal_contrast}

\begin{table*}[hbt!]
  \centering
  \small % Reduce font size to fit within column width
  \caption{Comparison of the baseline method with and without seasonal contrast (temporal views) when pre-training on SSL4EO, evaluated by the \gls{k-NN} protocol. The pre-training set is subsampled to \num{\sim 62500} locations that are present as four different timestamps.}
    
  \setlength{\tabcolsep}{6.8pt} % Adjust padding between columns
  \begin{tabular}{@{} l *{7}{S[table-format=2.2]} @{}}
  \toprule
  Method & {BEN-V2} & {EuroSAT} & {S4A-rand} & {S4A-tiles} & {So2Sat-rand} & {So2Sat-block} \\
  \midrule
  Seasonal Contrast      & 57.64 & \textbf{86.22} & 64.42 & 36.89 & 91.52 & \textbf{75.69} 	 \\
  No Seasonal Contrast   & \textbf{58.12} & 81.88 & \textbf{65.46} & \textbf{37.01} & \textbf{94.90} & 74.29 \\
  \bottomrule
  \end{tabular}
  \label{tab:seasonal_contrast}
\end{table*}

\begin{table*}[t!]
  \centering
  \small % Reduce font size to fit within column width
  \caption{Performance (in \%) of different center cropped image sizes for pre-training dataset SSL4EO with fixed downstream image resizing evaluated by the \gls{k-NN} protocol. The first image size (left of the arrow) is the center-cropped size of the pre-training dataset, and the second image size (right of the arrow) is the resized downstream image size.}
    
  \setlength{\tabcolsep}{5.2pt} % Adjust padding between columns
  \begin{tabular}{@{} l *{7}{S[table-format=2.2]} @{}}
  \toprule
  Image Size (Training Time) & {BEN-V2} & {EuroSAT} & {S4A-rand} & {S4A-tiles} & {So2Sat-rand} & {So2Sat-block} \\
  \midrule
  60x60\hspace{0.31cm} \textrightarrow 264x264  (\SI{3}{h}) & 57.15 & 84.01 & 65.01 & 36.06 & 94.66 & 73.74 \\
  120x120 \textrightarrow 264x264 (\SI{7}{h}) & \textbf{59.07} & \textbf{86.52} & \textbf{66.16} & 36.28 & \textbf{95.96} & \textbf{74.87} \\
  264x264 \textrightarrow 264x264 (\SI{25}{h})   & \textbf{59.22} & \textbf{86.70} & \textbf{66.15} & \textbf{36.67} & \textbf{95.85} & \textbf{74.91} \\
  \bottomrule
  \end{tabular}
  \label{tab:resize_pretraining}
\end{table*}

Seasonal contrast \cite{manas_seasonal_2021} is a contrastive \gls{SSL} approach tailored to \gls{RS}, which augments image diversity by incorporating different temporal instances of the same location in addition to standard augmentations. While conceptually compelling, prior evaluations of temporal views remain limited. Specifically, existing studies evaluate on a limited range of downstream tasks and suboptimal evaluation protocols (e.g., no k-NN evaluation). Moreover, the setup of Wang et al. \cite{wang_ssl4eo-s12_2023} implicitly favors seasonal contrast: although both models are trained on the same number of locations, the seasonal contrast model sees four times more images, as each location is represented across four time steps. To ensure a fair comparison, we equalize the image sets by treating all time steps as individual images when training without seasonal contrast. This ensures consistency in the image distribution across methods, isolating the effect of training strategy alone. Under this setup, we observe mixed results (\Cref{tab:seasonal_contrast}): while seasonal contrast benefits tasks such as EuroSAT and So2Sat-block, other tasks perform better when each time step is treated independently. We attribute this to the fact that contrastive \gls{SSL} in multispectral \gls{RS} images may implicitly encode temporal relationships through shared spatial and structural features. Explicitly enforcing temporal contrast can overconstrain the model in some cases, underscoring the need to tailor the integration of temporal views to specific downstream tasks.

\subsection{Image Size}
\label{sec:resizing}

Increasing image resolution at test time is known to improve performance for natural images \cite{richter_input_2021, touvron_fixing_2019} and has recently shown similar benefits for multispectral \gls{RS} tasks \cite{corley_revisiting_2024}. Independently, the \gls{CV} community has also established that higher training resolutions can enhance model performance, which likely motivated the use of $264 \times 264$ pixel images in multispectral \gls{RS} pre-training datasets such as SSL4EO and SeCo \cite{manas_seasonal_2021}. However, in multispectral \gls{RS}, larger image sizes only increase spatial coverage but not the spatial resolution, as this is determined by the satellite's onboard equipment, potentially leading to inefficient GPU utilization. We therefore evaluate pre-training with smaller image sizes ($120 \times 120$ and $60 \times 60$ center crops) compared to the standard $264 \times 264$ setup. Our results (\Cref{tab:resize_pretraining}) show that $120 \times 120$ images achieve equivalent downstream performance while reducing training time by a factor of three. The $60 \times 60$ variant performs slightly worse (\SI{2}{\percent}–\SI{4}{\percent}). Additional test-time resolution experiments with smaller image sizes, which confirm previously observed trends \cite{corley_revisiting_2024}, are provided in the supplemental material (\Cref{tab:resize_downstream_60}-\Cref{tab:resize_downstream_264}).

\section{Conclusion}
\label{sec:conclusion}
In this paper, we have introduced \textit{GeoRank}, a novel plug-in regularization term for contrastive \gls{SSL} that leverages geographical information to enhance representation alignment across geographically proximate images. By directly optimizing spherical distances, \textit{GeoRank} outperforms or matches prior methods that integrate geographical metadata and consistently improves diverse contrastive \gls{SSL} algorithms, demonstrating its generality as a framework-agnostic regularizer. Beyond this, we conducted a systematic study on key aspects of adapting contrastive \gls{SSL} for multispectral \gls{RS} imagery, offering practical insights into the design of the contrastive \gls{SSL} training pipeline. We show that adjusting the selection of data augmentation techniques to the unique properties of multispectral \gls{RS} imagery yields significant improvements. Our findings on temporal views provide a new perspective: while prior work suggests that different time steps as distinct views consistently improve performance, our experiments reveal that their effectiveness varies depending on the downstream task and dataset. Additionally, we challenge existing assumptions on dataset and image size, demonstrating that relatively smaller pre-training datasets and compact image sizes can yield strong performance on high-resolution multispectral data. This optimization offers substantial efficiency gains without compromising accuracy. Overall, our study highlights the necessity of tailoring contrastive \gls{SSL} methods to the distinct characteristics of multispectral \gls{RS} data, enabling more effective and efficient solutions for a wide range of applications in this domain. \vspace{0.2cm} \\
\noindent \textbf{Limitations.} The effectiveness of \textit{GeoRank}, and more generally of any method that integrates geographical information, relies on geographical overlap between the training and evaluation sets of the downstream task. When the downstream data exhibits a strong geographical domain shift (e.g., S4A-tiles), such that training and test regions are part of different countries, the benefits of geographic regularization diminish. In addition, \textit{GeoRank} is limited to contrastive \gls{SSL} methods, as it is formulated as a regularization term that builds on relationships between sample pairs. % Additionally, all experiments are primarily conducted on multispectral satellite images from the publicly available Sentinel-2 archive. The performances on other modalities or satellites (e.g., very high-resolution imagery) remains to be tested. 

{
    \small
    \bibliographystyle{ieeenat_fullname}
    \bibliography{main}
}

\newpage
\appendix
\clearpage

\section{Sentinel-2 Images and Data Processing Level}
\label{sec:sentinel2}

Sentinel-2 is a satellite mission from the European Space Agency (ESA), designed for Earth observation under the Copernicus program \cite{gascon_copernicus_2017}. It comprises two satellites, Sentinel-2A and Sentinel-2B, launched in 2015 and 2017, respectively. The mission provides high-resolution optical imagery for applications such as land cover classification, environmental monitoring, agricultural analysis, and emergency response. The Sentinel-2 satellites orbit the Earth in a sun-synchronous, polar orbit, capturing images of the entire planet approximately every five days. All Sentinel-2 data is freely accessible. 

\noindent \textbf{Spectral Bands and Spatial Resolution.} The \mbox{Sentinel-2} satellites carry a MultiSpectral Instrument (MSI) that captures optical images across 13 spectral bands, spanning from visible (RGB) and near-infrared (NIR) to short-wave infrared (SWIR) regions. These bands have different spatial resolutions, allowing for detailed analysis across diverse applications:
\begin{itemize}
\item \textbf{10 meters:} The four bands in this range (Blue, Green, Red, and NIR) are particularly useful for visual interpretations and land cover classifications due to their high resolution.
\item \textbf{20 meters:} Six bands fall in this range, including red edge and short-wave infrared bands, which are instrumental in vegetation analysis, water quality monitoring, and distinguishing various land cover types.
\item \textbf{60 meters:} Three bands in this range are primarily used for atmospheric correction and cloud screening, with a coarser resolution that provides broader spatial coverage rather than detailed surface features.
\end{itemize}

\noindent \textbf{Data Processing Levels.} Further, Sentinel-2 images are provided at two processing levels, tailored to meet different user needs:
\begin{itemize}
\item \textbf{Level-1C (L1C):} L1C data consists of Top-Of-Atmosphere (TOA) reflectance values, meaning it captures reflectance as observed from the satellite. This processing level includes the effects of atmospheric conditions like haze and scattering, making it ideal for users who perform their own atmospheric corrections.
\item \textbf{Level-2A (L2A):} L2A data provides Bottom-Of-Atmosphere (BOA) reflectance values, which means atmospheric corrections have been applied to adjust for atmospheric interference. This data is ready for immediate analysis, allowing users to focus on surface-level characteristics without needing to handle atmospheric correction.
\end{itemize}

\section{Dataset Details}
\label{sec:dataset_details}

All Sentinel-2-based multispectral datasets used in this paper are preprocessed to L2A processing. Out of the originally \num{13} bands, \num{10} bands with a spatial resolution of either \SI{10}{\metre} or \SI{20}{\metre} are selected for experiments. In the following, the pre-training and downstream datasets are described in detail:  \vspace{0.2cm} \\
% \noindent \textbf{SeCo} \cite{manas_seasonal_2021} is a large-scale unlabeled dataset designed to support self-supervised learning (SSL) in remote sensing. It consists of around \num{1000000} unlabeled images of size $264 \times 264$ pixels from across diverse geographic regions of the Earth. Each of the \num{200000} image locations is represented with five different timestamps (one per season, two in Summer). The SeCo sampling procedure involves uniformly selecting coordinates within a \SI{50}{\km} radius of the \num{10000} most populated cities followed by a selection of five random dates between Summer 2019 and Summer 2020 (one per season within a 15-day interval). The time series is included in the dataset if each timestamp has at least one tile with less than \SI{10}{\percent} cloud coverage across all seasonal intervals. It is worth noting that the sampling strategy has a strong bias towards populated regions in the northern hemisphere. Locations around the equator are less likely to be selected due to persistent cloud cover throughout the year. \vspace{0.2cm} \\ 
\noindent \textbf{SSL4EO} \cite{wang_ssl4eo-s12_2023} is a large-scale unlabeled dataset designed to support \gls{SSL} in remote sensing. It consists of images of size $264 \times 264$ pixels from approximately \num{250 000} diverse locations on Earth that are represented by four seasonal timestamps within the years 2020 and 2021. A time series is included in the dataset if each seasonal interval of 90 days contains at least one tile with less than \SI{10}{\percent} cloud coverage. SSL4EO builds upon the sampling strategy of SeCo \cite{manas_seasonal_2021}, which selected multi-seasonal image time series within a \SI{50}{\km} radius of the \num{10000} most populated cities worldwide. To address the spatial redundancy introduced by this approach (oversampling), SSL4EO enforces non-overlapping geographic coverage across image locations. In addition to the Sentinel-2 images in L2A processing, each image is further associated with the same image in L1C processing and an image acquired by the radar satellite Sentinel-1. In this paper, we only use the Sentinel-2 image in L2A processing. It is worth noting that the sampling strategy has a strong bias towards populated regions in the northern hemisphere. Locations around the equator are less likely to be selected due to persistent cloud cover throughout the year. \vspace{0.2cm} \\
\noindent \textbf{EuroSAT} \cite{helber_eurosat_2019} is a single-label classification dataset with \num{27000} labeled images of size $64 \times 64$. It is annotated with \num{10} land-cover classes that include categories such as forests, agricultural areas, water bodies, and urban zones. The class annotations are derived from the European Urban Atlas. We utilize a stratified train/test/validation split that is composed of \SI{60}{\percent}, \SI{20}{\percent}, \SI{20}{\percent}, respectively. The original version of the dataset is published in L1C processing. To standardize all datasets we converted the images to the L2A processing, and denote the processed dataset as EuroSAT-L2A. \vspace{0.2cm} \\
\noindent \textbf{So2Sat} \cite{zhu_so2sat_2020} is a single-label classification dataset with approximately \num{400000} labeled image pairs from the satellites Sentinel-1 (radar) and Sentinel-2 (optical) acquired over \num{50} metropolitan areas worldwide. Each image is of size $32 \times 32$. The \num{17} classes capture both urban and non-urban land cover types and are derived from OpenStreetMap (OSM) data. The dataset provides three different splits to evaluate model performance under varying conditions. The random split (So2Sat-random) divides images randomly across training and test sets (\SI{80}{\percent}, \SI{20}{\percent}). The block split (So2Sat-block) partitions data based on geographically distinct but neighboring blocks, ensuring less correlation between training and test images (\SI{80}{\percent}, \SI{20}{\percent}).%  The culture10 split (So2Sat-culture) groups the metropolitan areas into a training set (40 areas) and a test set (10 areas), ensuring a geographical domain shift between both sets.
For standardized experiments, we selected only the Sentinel-2 images.\vspace{0.2cm} \\
\noindent \textbf{BigEarthNet-V2 (BEN-V2)} \cite{clasen_reben_2024} is a refined version of the large-scale multi-label dataset BigEarthNet-S2 \cite{sumbul_bigearthnet_2019} that includes \num{590326} images acquired over ten countries in Europe. Each image is of size $120 \times 120$. The land use land cover (LULC) class annotations are obtained from the CLC inventory \cite{buttner_corine_2004}. Following the LULC class nomenclature proposed in \cite{sumbul_bigearthnet-mm_2021}, each image is annotated with a subset of $19$ LULC classes, including different types of forests, water, or complex urban or agricultural classes. We utilize a filtered subset that excludes images with seasonal snow, clouds, and cloud shadows. The selected subset is divided by a block-wise split into a training set (\SI{50}{\percent}), a validation set (\SI{25}{\percent}), and a test set (\SI{25}{\percent}). Each set can contain different timestamps of the same geographical location.  \vspace{0.2cm} \\
\noindent \textbf{Sen4Agri-ML} is a multi-label classification dataset that was created based on the semantic segmentation dataset Sen4AgriNet \cite{sykas_sen4agrinet_2021} designed for agricultural monitoring. All images are acquired over France and Catalonia in the years 2019 and 2020. The originally \num{225000} images of size $366 \times 366$ composed as time series data were subsampled into images of size $120 \times 120$. For each time series, one representative image in the summer months was randomly selected. The respective multi-labels were derived from the $120\times120$-pixel segmentation maps. Further, all images containing no class were discarded. The \num{9} high-level crop type class annotations originate from farmer declarations collected via the Land Parcel Identification System (LPIS) \cite{owen_land_2016}.  We utilize the random train/test split (denoted as S4A-random) and the tiles-based train/test split (denoted as S4A-tiles) that is composed of training images from France in 2019 and test images from Catalonia in 2020.\vspace{0.2cm} \\
\noindent \textbf{CashewPlant} \cite{lacoste_geo-bench_2023} is a semantic segmentation dataset derived from Sentinel-2 imagery collected over approximately \SI{120}{\kilo\meter\squared} in central Benin. It consists of images of size $256 \times 256$. Each image is annotated with pixel-wise masks that distinguish seven classes: well-managed plantations, poorly managed plantations, non-plantation, residential areas, background, uncertain, and no-data. The annotations were generated from field surveys with handheld GPS devices and refined with very high-resolution Pléiades imagery. In the GEO-Bench version, the dataset is divided into training (\SI{75}{\percent}), validation (\SI{20}{\percent}), and test (\SI{5}{\percent}) splits.

\begin{figure}[]
    \centering
    \def\segnoisefrac{.24}
    \begin{subfigure}{\segnoisefrac\linewidth}
        {\includegraphics[width=\linewidth]{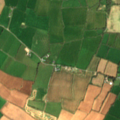}}
    \end{subfigure}
    \begin{subfigure}{\segnoisefrac\linewidth}
        {\includegraphics[width=\linewidth]{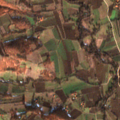}}
    \end{subfigure}
    \begin{subfigure}{\segnoisefrac\linewidth}
        {\includegraphics[width=\linewidth]{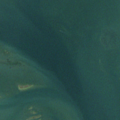}}
    \end{subfigure}
    \begin{subfigure}{\segnoisefrac\linewidth}
        {\includegraphics[width=\linewidth]{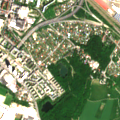}}
    \end{subfigure} \\ \vspace{0.05cm}
    \begin{subfigure}{\segnoisefrac\linewidth}
        {\includegraphics[width=\linewidth]{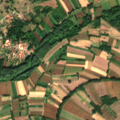}}
    \end{subfigure}
    \begin{subfigure}{\segnoisefrac\linewidth}
        {\includegraphics[width=\linewidth]{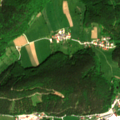}}
    \end{subfigure}
    \begin{subfigure}{\segnoisefrac\linewidth}
        {\includegraphics[width=\linewidth]{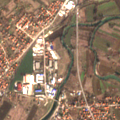}}
    \end{subfigure}
    \begin{subfigure}{\segnoisefrac\linewidth}
        {\includegraphics[width=\linewidth]{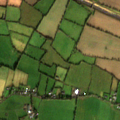}}
    \end{subfigure}
    \caption{Example Sentinel-2 images taken from BEN-V2.}
    \label{fig:example_sentinel2_images}
\end{figure}

\section{Implementation Details}
\label{sec:implementation_details}

This section in detail describes the hyperparameters used to train and evaluate the models.

\subsection{Data Preprocessing}
The reflectance values captured by Sentinel-2 are stored in an \textit{uint16} format. However, the distribution of values is highly skewed towards values within the range of \num{0} to \num{4000}, with a long tail distribution reaching values up to $2^{13}$. To be able to apply channel augmentation techniques to \mbox{Sentinel-2} data, we preprocess the \textit{uint16} values to \textit{uint8} values by dividing each channel by its $99^{\text{th}}$ percentile for BEN-V2, So2Sat, Sen4Argi-ML and EuroSAT-L2A, followed by a 0-1-clipping and a multiplication by 255. For SSL4EO we divide each channel by its $95^{\text{th}}$ percentile due to a larger long tail in the distribution since both pre-training datasets comprise a higher fraction of images with partial cloud cover. The exact values for the percentiles for each channel can be found in the code repository published together with this paper.

\subsection{General Hyperparameter}
All self-supervised methods are implemented via packages lightning \cite{falcon_pytorch_2019} and lightly \cite{susmelj_lightly_2020}. For both the contrastive self-supervised pre-training and the three downstream evaluation protocols we set the batch size to \num{512}. For contrastive self-supervised pre-training that involve MoCoV2 \cite{chen_improved_2020}, we use the LARS optimizer with a learning rate of \num{0.4}, momentum of \num{0.9} and a weight decay of \num{0.000001} and train the network for \num{50} epochs. The model with the lowest training loss is selected for downstream evaluation. The InfoNCE is applied with a memory bank size of \num{4092} and the temperature value of \num{0.04}. For contrastive self-supervised pre-training with SimCLR \cite{chen_simple_2020}, BYOL \cite{grill_bootstrap_2020} and SimSiam \cite{chen_exploring_2021} we use an SGD optimizer with learning rate of \num{0.06}. The NT-Xent loss for SimCLR follows the default setup with a temperature value of \num{0.5}. BYOL and SimSiam are trained with negative cosine similarity. For DINO \cite{caron_emerging_2021} pre-training we use an Adam optimizer with learning rate of \num{0.001}. The momentum of the exponential moving average of the model for MoCoV2, BYOL and DINO is compute by a cosine schedule via \num{10} steps from \num{0.996} to \num{1}. The DINO loss has an output dimension of \num{2048} and epochs for the teacher temperature warmup is set to \num{5}. The rest follows the default hyperparameter setting of lightly. For DINO we employ two local views at size $60 \times 60$ with scale factor for RRC of (\num{0.25}, \num{0.5}) and two global views at size $120 \times 120$ with scale factor for RRC of (\num{0.5}, \num{1.0}). Similar to MoCoV2, all models are trained for \num{50} epochs. For \textit{GeoRank} we use the hyperparamter $\alpha$ set to \num{0.48} and $d_{\text{max}}$ set to \num{2500}. The set of data augmentation techniques used for pre-training includes RRC with a ratio of (\num{0.75}, \num{1.33}) and a scale of (\num{0.2}, \num{1.0}) applied with a probability of \num{1.0}, a flip operation (horizontally and vertically) applied with a probability of \num{0.75} and RR90 applied with a probability of \num{0.75}. For the differentiable softrank function that we use to approximate of the rank function we use regularization strength of \num{0.001} and perform l2 regularization. For MAE training, we adopt the default hyperparameters proposed in the original paper. All three RS MAE variants are trained with a batch size of \num{16}, and learning rate scheduling is deactivated. A masking ratio of \num{0.75} is used, along with \num{10} warm-up epochs and the AdamW optimizer (with betas set to (\num{0.9}, \num{0.95}) for SatMAE \cite{cong_satmae_2022} and ScaleMAE \cite{reed_scale-mae_2023}). For SatMAE, the output size of the random resized crop (RRC) is set to $96 \times 96$, with a scale range of (\num{0.6}, \num{1.0}). Feature extraction is performed using all tokens except the class token. The weight decay is set to \num{0.0}, and the learning rate is \num{0.0001}. For CrossScaleMAE \cite{tang_cross-scale_2023} and ScaleMAE, the RRC output size is set to $112 \times 112$. Additionally, ScaleMAE internally maintains a target size of $224 \times 224$ using a constant source size scheduler. Both models use a weight decay of \num{0.05}. The learning rate is set to \num{0.00005} for CrossScaleMAE and \num{0.00015} for ScaleMAE. In analogy to contrastive methods the maximum training epochs are set to \num{50}. The only data augmentation used in downstream training is random flipping with a probability of \num{0.8}. To save computational cost, the standard preprocessing of SSL4EO consists of a $120 \times 120$ pixels centre crop (except for \Cref{sec:resizing}) and a training set that consists of one randomly selected timestamp per location (except for \Cref{sec:seasonal_contrast}). The pre-training set for the experiments with temporal views (see \Cref{sec:seasonal_contrast}) is subsampled to \num{\sim 62500} locations with each location being present with four different timestamps to avoid measuring artifacts of pre-training dataset saturation.

\subsection{Evaluation Protocols}
The k-NN evaluation protocol applies a k-NN clustering to the learned representations, the linear evaluation protocol freezes the model backbone and trains a simple linear layer on top of the learned representations, while the fine-tuning protocol re-trains all layers of the backbone. For k-NN evaluation, we set the number of clusters to \num{10} and the sharpening parameter to \num{0.9}. For linear evaluation and fine-tuning we train for \num{30} epochs with an AdamW optimizer scheduled by a cosine annealing learning rate scheduling with a start rate set to \num{0.001} and warm-up iterations based on the number of steps. The weight decay is set to \num{0.01}. Supervised training from scratch is conducted with the same hyperparameter setting as the fine-tuning evaluation protocol. The evaluation protocol for semantic segmentation tasks employs a UPerNet decoder \cite{xiao_unified_2018} that receives frozen features from layer 1 to 4 for ResNet backbones and is trained for \num{50} epochs. Hidden feature size is set to \num{256} and and output feature size is set to \num{128}. We train the UPerNet with an SGD optimization with learning rate \num{0.02}, momentum \num{0.9} and weight decay of \num{0.0001}. For transformer backbones, we construct multi-scale feature maps by reshaping the encoder sequence into grids of shape $(\tilde{c}, h^\prime, w^\prime)$ at resolutions $1/4$, $1/8$, $1/16$, and $1/32$. For CrossScaleMAE, the final pyramid level is handled separately by unfolding and bilinearly interpolating features to the target resolution. Channel dimensions are reduced via group-wise averaging to match UPerNet’s expected inputs ($64$, $128$, $256$, $512$). This process, applied independently to each quarter of the transformer blocks, yields a four-level pyramid compatible with standard convolutional decoders.

\subsection{Data Augmentation}
The default data augmentation pipeline adopted from computer vision (CV) includes RRC with the same hyperparameter setting as in the general hyperparameter, ColorJitter (only applying Contrast and Brightness adjustments) with a limit of \num{0.4} applied with a probability of \num{0.8}, GrayScale applied with a probability of \num{0.2}, GaussianBlur with a sigma of (\num{0.1}, \num{2.0}) applied with a probability of \num{0.5} and horizontal flipping applied with a probability of \num{0.5}. Hue and Saturation are not defined for more than 3 channels. For the ablation study, we add individual augmentation techniques to the three geometric augmentation techniques from the general hyperparameter with a probability of \num{0.2}. The base magnitudes can be seen in the right column of \Cref{tab:augmentation_ablation}. If applicable these are applied with a scalar of \num{1}, \num{2} or \num{3}. We resized the datasets So2Sat-rand and So2Sat-block to a spatial resolution of $120 \times 120$ pixels for all experiments except for \Cref{sec:resizing}. All augmentation techniques are taken from the \texttt{albumentation} library.

\subsection{Compute Resources}
All experiments were conducted on an internal server equipped with 2× AMD EPYC 9554 64-core processors (256 threads), 6× NVIDIA H100 PCIe GPUs (each with 81 GB memory, CUDA 12.2), and 1.5 TiB of system RAM. The system runs Ubuntu 22.04 with Linux kernel 5.15 and NVIDIA driver version 535.183.01. Each training run was executed on a dedicated GPU. Standard pre-training took between 4 and 7 hours, depending on the dataset size and the extent of data augmentation. K-Nearest Neighbors evaluations for downstream tasks required up to 15 minutes per dataset, while fine-tuning evaluations took up to 3 hours. The pre-training of experiments involving geographical regularization required between 10 and 14 hours. Moreover, reproducing results from Ayush et al. involved precomputing k-means clusters, which incurred an additional small overhead.

\section{Extended Experiments}
\label{sec:extended_experiments}

In this section, we present complementary results on different pre-training datasets.

\subsection{Data Augmentation Ablation for Geometric Augmentation}
\label{sec:dataaug_ablation_geometric}

We extend the ablation study presented in \Cref{sec:dataaug} and also evaluate the average performance on all downstream tasks on permutations of the three geometric augmentation techniques \gls{RRC}, Flip and \gls{RR90} (see \Cref{tab:geometric_ablation}). The results indicate that the biggest driver for downstream performance is \gls{RRC}. Nonetheless, the combinations of \gls{RRC} with \gls{RR90} and Flip and  yield the highest averaged downstream performance.

\begin{table}
  \centering
  \small
  \caption{Ablation study for enabling or disabling one of the three basic geometric augmentation techniques. Performance is the averaged score (Avg. Result) in the \gls{k-NN} protocol over all six downstream tasks when pre-training on SSL4EO.}
  \vspace{0.2cm}
   \begin{tabular}{@{} c c c c @{}} 
    \toprule
    RRC & RR90 & Flip & Avg. Result \\
    \midrule
    \checkmark & - & - & 63.64 \\
    - & \checkmark & - & 60.19 \\
    - & - & \checkmark & 55.07 \\
    \checkmark & \checkmark & - & \textbf{68.59} \\
    \checkmark & - & \checkmark & 65.73 \\
    - & \checkmark & \checkmark & 62.25 \\
    \midrule
    \checkmark & \checkmark & \checkmark & \textbf{68.62}\\
    \bottomrule
  \end{tabular}
  \label{tab:geometric_ablation}
\end{table}

\subsection{Data Augmentation for BEN-V2}
\label{sec:dataaug_benv2_seco}

\begin{figure*}[ht!]
    \centering
    \def\segnoisefrac{.32}
    \begin{subfigure}{\segnoisefrac\linewidth}
        {\includegraphics[width=\linewidth]{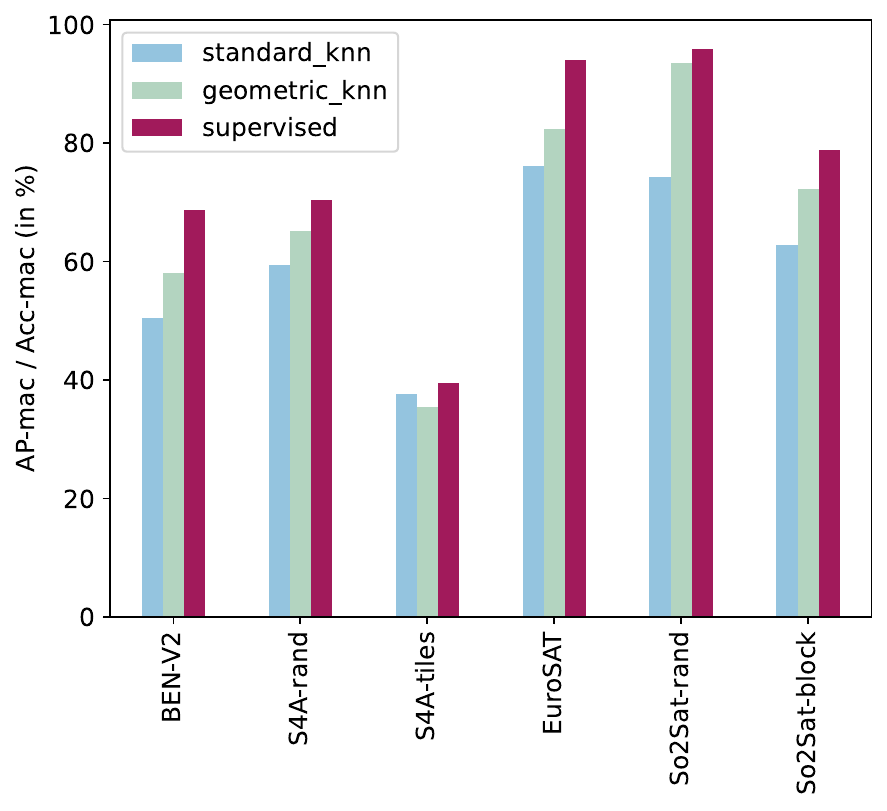}}
        {\caption{}\label{subfigure:benv2_a}}
    \end{subfigure}
    \begin{subfigure}{\segnoisefrac\linewidth}
        {\includegraphics[width=\linewidth]{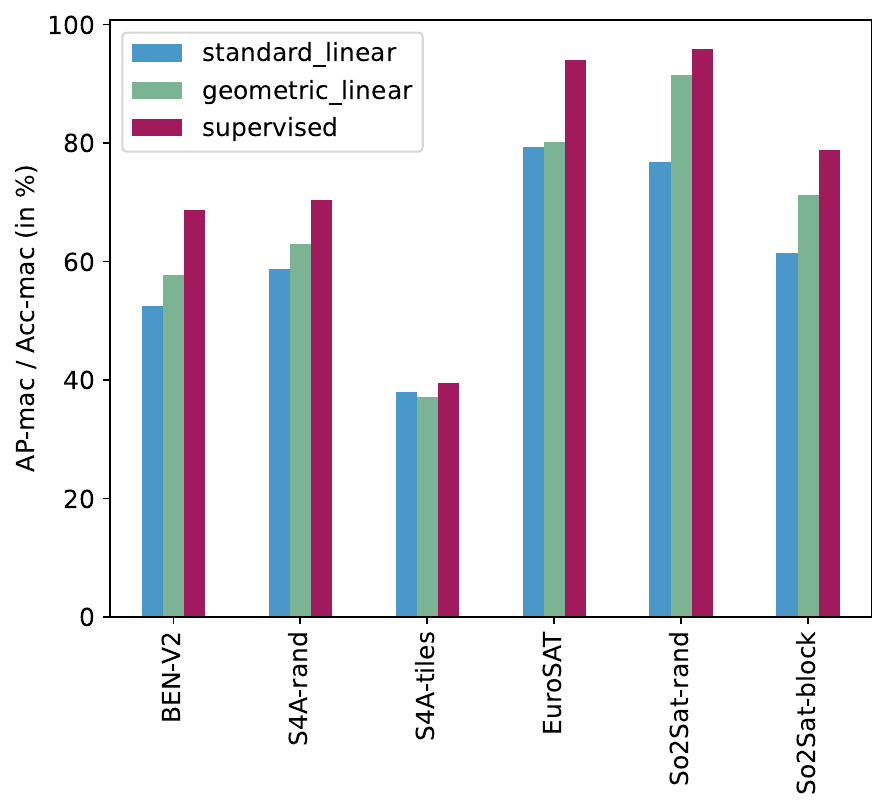}}
        {\caption{}\label{subfigure:benv2_b}}
    \end{subfigure}
    \begin{subfigure}{\segnoisefrac\linewidth}
        {\includegraphics[width=\linewidth]{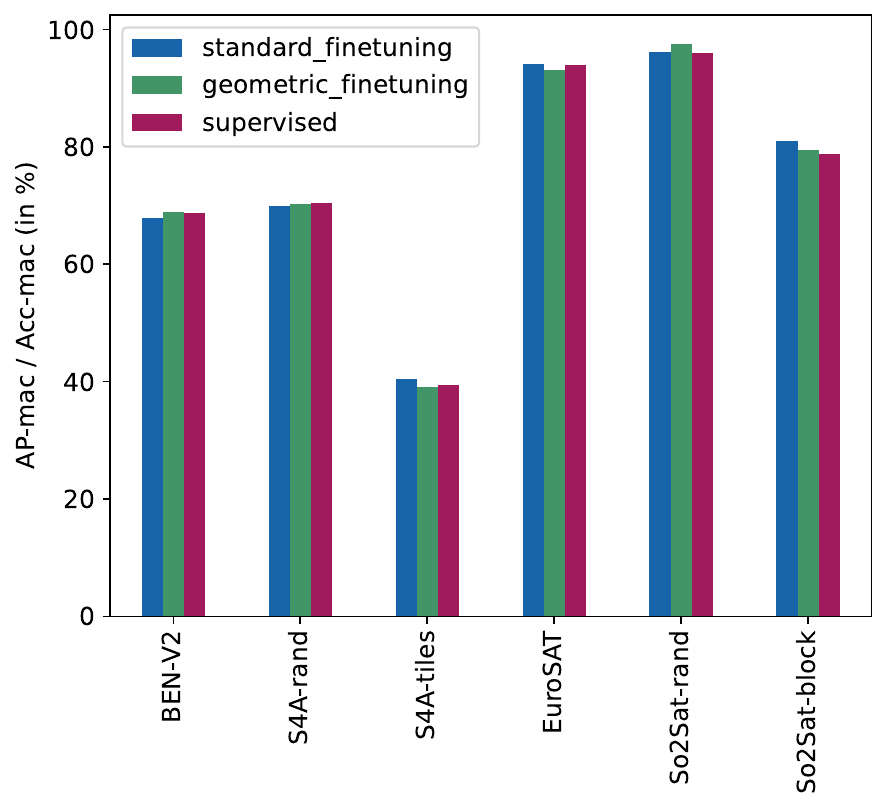}}
        {\caption{}\label{subfigure:benv2_c}}
    \end{subfigure}
    \caption{Performance comparison between the standard augmentation pipeline (blue), the geometric augmentation pipeline (green) and supervised training (red) on all six downstream tasks when pre-training on BEN-V2: (a) evaluated with the \gls{k-NN} evaluation protocol, (b) evaluated with the linear evaluation protocol, (c) evaluated with the fine-tuning protocol.}
    \label{fig:dataaug_comparison_benv2}
\end{figure*}

In line with the results of comparing the default computer vision data augmentation pipeline with a geometric augmentation pipeline when pre-training on the SSL4EO dataset, we find that the geometric pipeline outperforms the standard pipeline by values of up to \SI{15}{\percent} in the k-NN protocol when pre-training on BEN-V2 (see \Cref{subfigure:benv2_a}). It is noteworthy that this effect diminishes when more hyperparameters are involved in the evaluation protocol: while the average improvement for linear evaluation is between \SI{3}{\percent} and \SI{5}{\percent}, the differences become marginal when observing the results for the fine-tuning protocol (see \Cref{subfigure:benv2_c}). Especially the evaluation under the k-NN protocol emphasizes the relevance of adjusting the data augmentation pipeline to multispectral RS images.

\subsection{Qualitative Analysis of Representation Space}
\label{sec:analysis_representation_space}

To assess the qualitative effect of the proposed regularization, we compare latent representations obtained from the baseline model (MoCoV2) and MoCoV2 with \textit{GeoRank}. From the BEN-V2 training set, we randomly sample 2560 images and extract features from the penultimate layer of each model. The resulting representations are reduced in dimensionality using principal component analysis (PCA) to 50 components, followed by t-SNE with perplexity set to 30 and learning rate set to 200. The first row of Figure~X visualizes the embeddings colored by normalized latitude, while the second row uses normalized longitude. MoCoV2 with \textit{GeoRank} exhibits smoother spatial organization in the embedding space, with representations reflecting relative geographical ordering rather than forming rigid clusters. This observation is consistent with the intended rank-based formulation, which preserves ordering relations without enforcing strict alignment.

\begin{figure*}[t!]
    \centering
    \def\segnoisefrac{.4}
    \begin{subfigure}{\segnoisefrac\linewidth}
        {\includegraphics[width=\linewidth]{images/tsne_georank_latitude.pdf}}
    {\caption{}\label{subfigure:georank_latitude}}
    \end{subfigure}
    \hspace{0.8cm}
    \begin{subfigure}{\segnoisefrac\linewidth}
        {\includegraphics[width=\linewidth]{images/tsne_baseline_latitude.pdf}}
        {\caption{}\label{subfigure:baseline_latitude}}
    \end{subfigure} \\
    \begin{subfigure}{\segnoisefrac\linewidth}
        {\includegraphics[width=\linewidth]{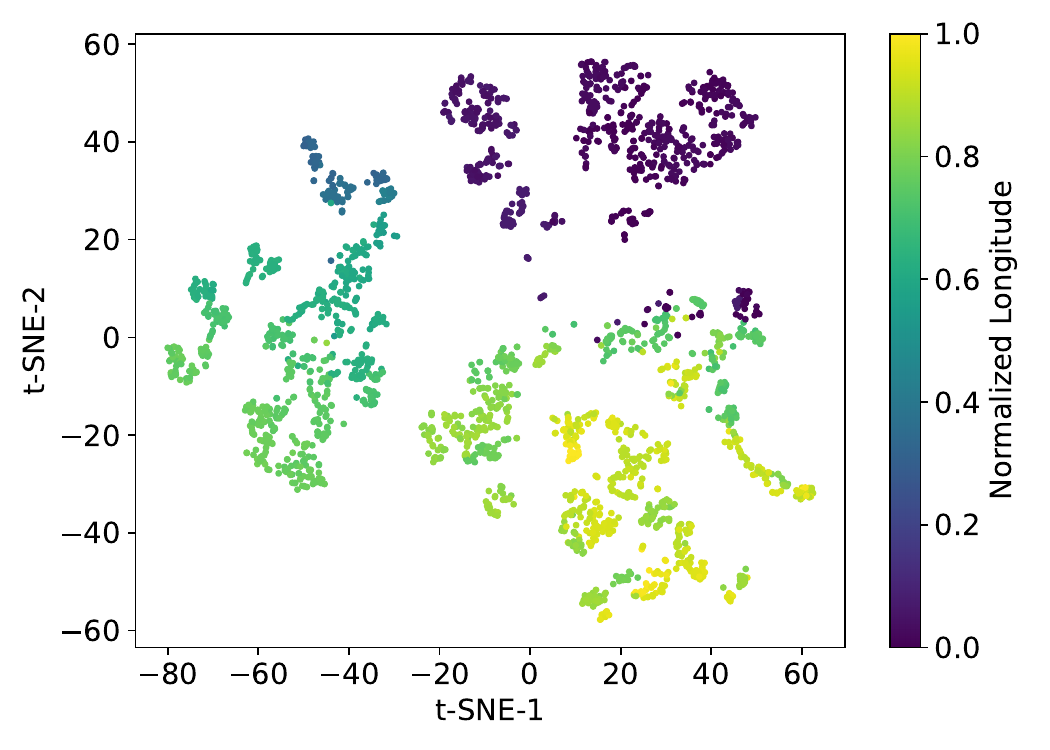}}
        {\caption{}\label{subfigure:georank_longitude}}
    \end{subfigure}
    \hspace{0.8cm}
    \begin{subfigure}{\segnoisefrac\linewidth}
        {\includegraphics[width=\linewidth]{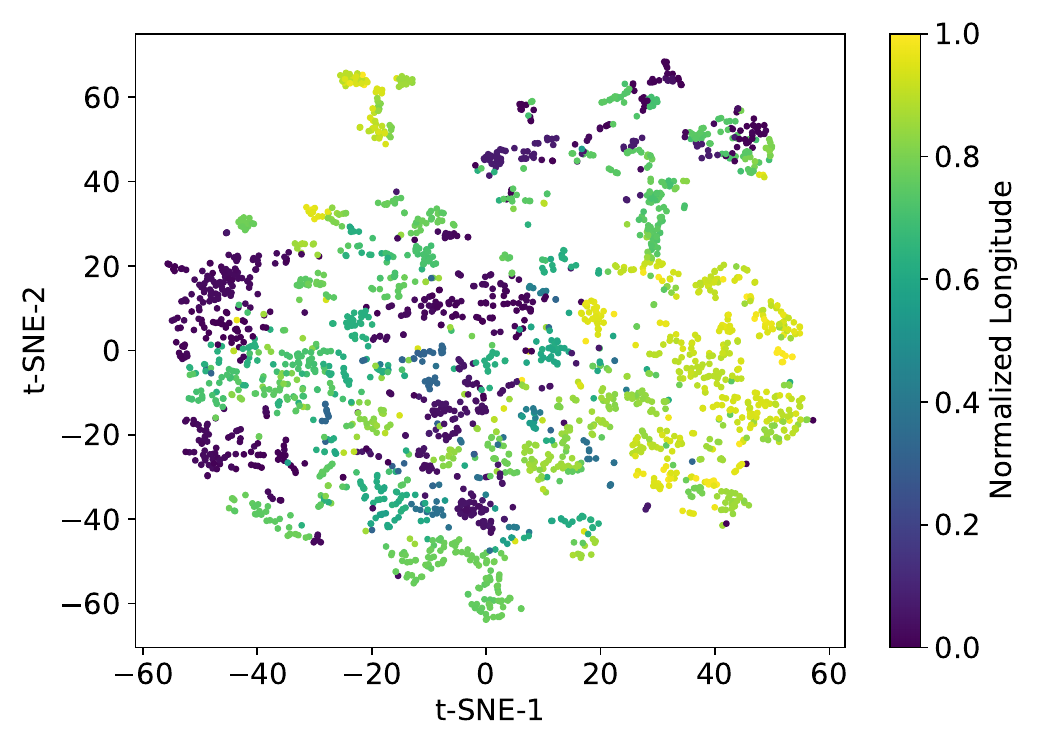}}
        {\caption{}\label{subfigure:baseline_longitude}}
    \end{subfigure}
    \caption{t-SNE of penultimate layer representations for 2560 BENV2 samples after PCA (50 components). Points are colored by normalized latitude in (a) MoCoV2 with \textit{GeoRank} and (b) MoCoV2, and by normalized longitude in (c) MoCoV2 with \textit{GeoRank} and (d) MoCoV2.}
    \label{fig:t-SNE_analysis_full}
\end{figure*}

\subsection{Compatibility with RS-Specific Contrastive SSL Methods}

Beyond standard contrastive algorithms, we also tested \textit{GeoRank} with \gls{RS}-specific \gls{SSL} methods that incorporate temporal or multimodal information. Specifically, we combined \textit{GeoRank} with Seasonal Contrast (SeCo) \cite{manas_seasonal_2021} and CROMA \cite{fuller_croma_2023}, which represent temporal and multimodal contrastive learning respectively. Results are reported in \Cref{tab:extending_rs_ssl_methods}. Improvements are generally modest, with gains on most benchmarks. Consistent with the main experiments, \textit{GeoRank} shows a drop in performance only in the presence of geographical domain shift, as observed on S4A-tiles. Overall, these experiments confirm that \textit{GeoRank} can be integrated into temporal and multimodal SSL setups without interfering with their design objectives.

\begin{table*}[t!]
  \centering
  \small % Reduce font size to fit within column width
  \caption{Extending existing RS-specific SSL methods with \textit{GeoRank} , when pre-training on SSL4EO, evaluated by the \gls{k-NN} protocol.}
    
  \setlength{\tabcolsep}{6.8pt} % Adjust padding between columns
  \begin{tabular}{@{} l c c c c c c c @{}}
  \toprule
  Method & {BEN-V2} & {EuroSAT} & {S4A-rand} & {S4A-tiles} & {So2Sat-rand} & {So2Sat-block} & {BEN-V2 S1+S2}\\
  \midrule
  CROMA \cite{fuller_croma_2023} & 61.13 & 89.77 & 65.53 & \textbf{36.23} & 94.69 & \textbf{75.09} & 61.62 \\
  % CROMA \cite{fuller_croma_2023} & 61.13 & \textbf{90.46} & 65.53 & \textbf{36.23} & 94.69 & \textbf{75.09} & 61.65 \\
  CROMA \cite{fuller_croma_2023} + GeoRank & \textbf{61.34} & \textbf{89.96} & \textbf{65.67} & 35.27 & \textbf{94.79} & 7\textbf{5.10 }& \textbf{61.72} \\
  \midrule
  SeCo \cite{manas_seasonal_2021}     & 57.64 & 86.22 & 64.42 & \textbf{36.89} & 91.52 & \textbf{75.69} & \\
  SeCo \cite{manas_seasonal_2021} + GeoRank & \textbf{57.99} & \textbf{86.60} & \textbf{64.72} & 36.34 & \textbf{92.08} & \textbf{75.67} & - \\
  \bottomrule
  \end{tabular}
  \label{tab:extending_rs_ssl_methods}
\end{table*}

\subsection{Dataset Cardinality for SSL4EO under different Model Sizes}
\label{sec:dataset_size_model_sizes}

\begin{figure*}[t!]
    \centering
    \def\segnoisefrac{.4}
    \begin{subfigure}{\segnoisefrac\linewidth}
        {\includegraphics[width=\linewidth]{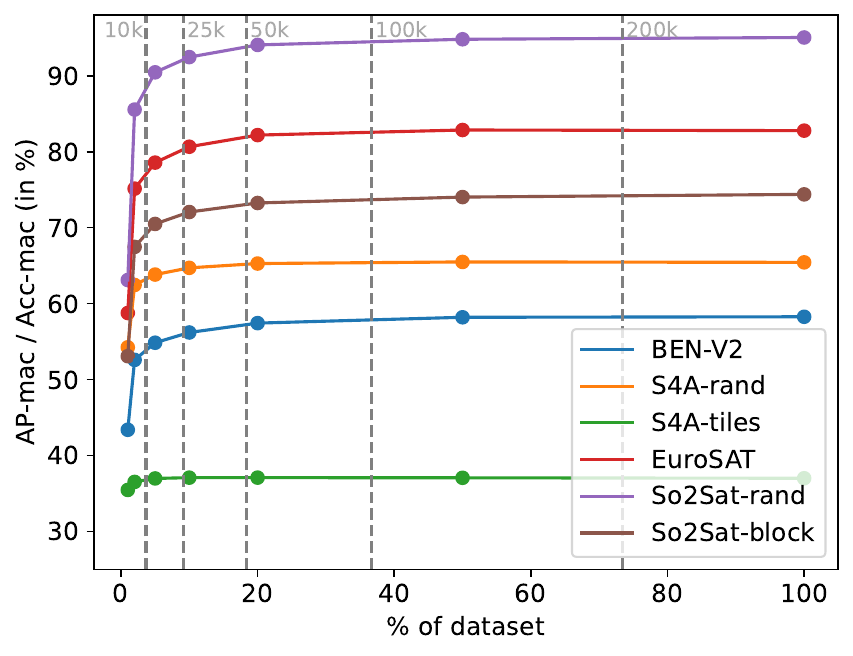}}
        {\caption{}\label{subfigure:pretraining_size_ssl4eo_resnet18}}
    \end{subfigure}
    \hspace{0.8cm}
    \begin{subfigure}{\segnoisefrac\linewidth}
        {\includegraphics[width=\linewidth]{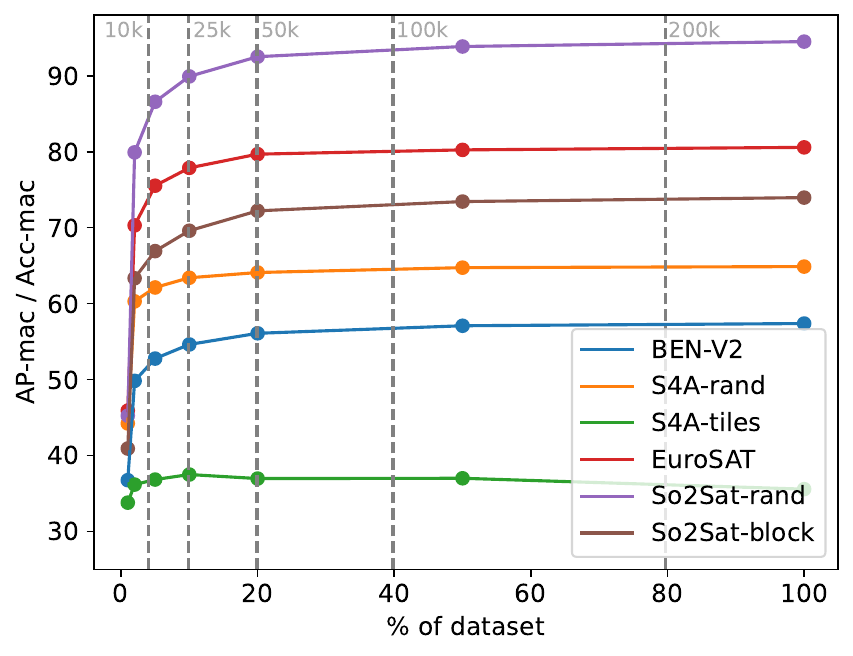}}
        {\caption{}\label{subfigure:pretraining_size_ssl4eo_resnet34}}
    \end{subfigure} \\
    \begin{subfigure}{\segnoisefrac\linewidth}
        {\includegraphics[width=\linewidth]{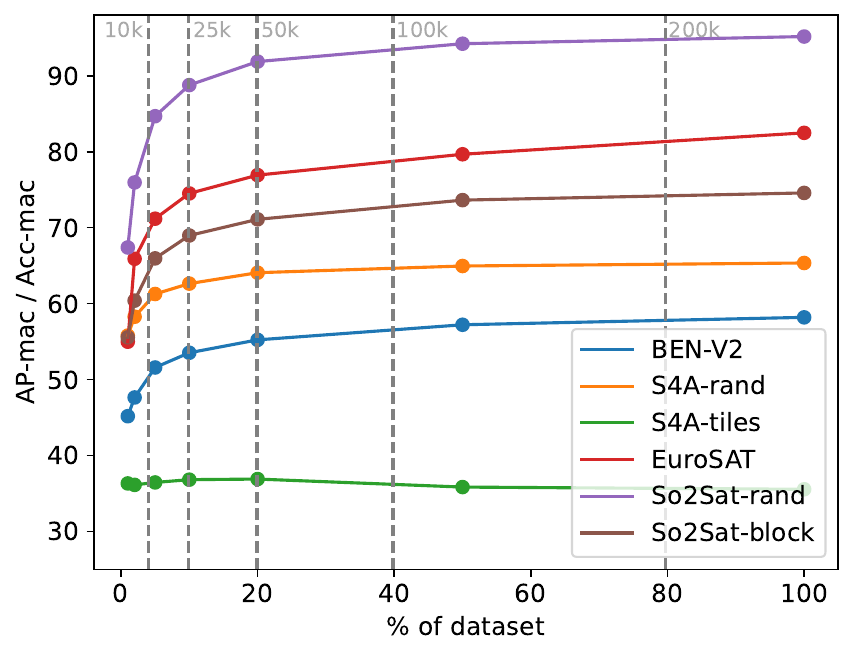}}
        {\caption{}\label{subfigure:pretraining_size_ssl4eo_resnet50}}
    \end{subfigure}
    \hspace{0.8cm}
    \begin{subfigure}{\segnoisefrac\linewidth}
        {\includegraphics[width=\linewidth]{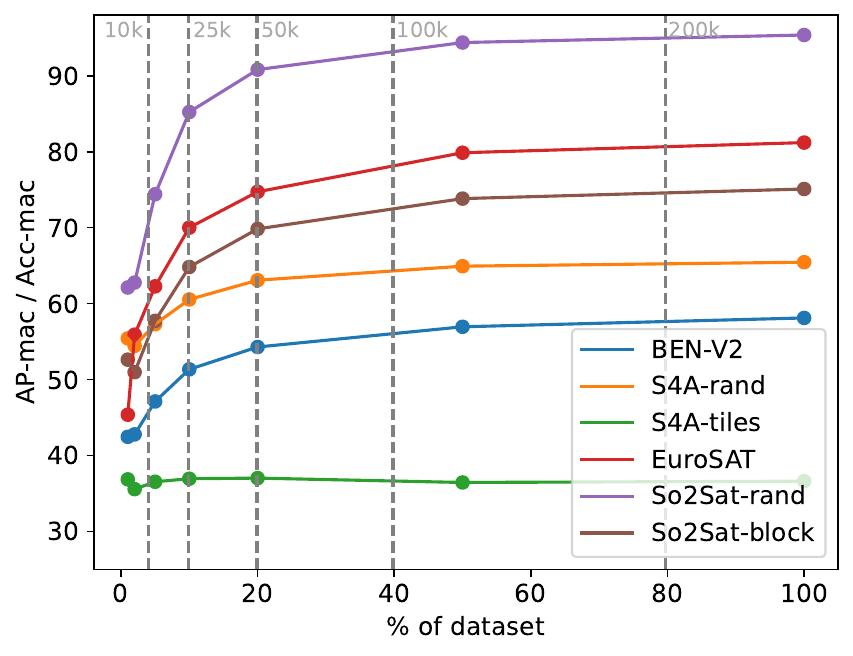}}
        {\caption{}\label{subfigure:pretraining_size_ssl4eo_resnet101}}
    \end{subfigure}
    \caption{Performance of different subset sizes of the pre-training dataset SSL4EO evaluated on all six downstream tasks by \gls{k-NN} with different backbones. (a) ResNet18. (b) ResNet34. (c) ResNet50. (d) ResNet101.}
    \label{fig:pretraining_size_model_size}
\end{figure*}

Against the hypothesis of Wang et al. \cite{wang_ssl4eo-s12_2023}, we observe no significant differences in saturation for different model sizes when we pre-train different sizes of ResNets on SSL4EO (see \Cref{fig:pretraining_size_model_size}).

\subsection{Downstream Image Size for different Pre-Training Image Sizes}
\label{sec:image_size_extended}

We find that for a fixed pre-training image size of $60 \times 60$ pixels, $120 \times 120$ pixels or $264 \times 264$ pixels, resizing the downstream images to larger image sizes tends to result in an increase in performance for all downstream tasks (see \Cref{tab:resize_downstream_60}, \Cref{tab:resize_downstream_120} and \Cref{tab:resize_downstream_264}). Similar to Corley et al. \cite{corley_revisiting_2024}, we observe a saturation effect at $264 \times 264$ pixels for the resizing of the downstream task for a pre-training image size of $264 \times 264$ pixels. We note that for a larger gap between pre-training image size and downstream resizing, e.g., $60 \times 60$ to $264 \times 264$, a downstream resizing of $120 \times 120$ can be already effective.

\begin{table*}[hbt!]
  \centering
  \small % Reduce font size to fit within column width
  \caption{Performance (in \%) of different resizing strategies for downstream datasets evaluated by the \gls{k-NN} protocol when pre-training on SSL4EO with fixed image size. The first image size (left of the arrow) is the center cropped size of the pre-training dataset, and the second image size (right of the arrow) is the resized downstream image size.}
  \setlength{\tabcolsep}{7pt} % Adjust padding between columns
  \begin{tabular}{@{} l *{7}{S[table-format=2.2]} @{}}
  \toprule
  Image Size & {BEN-V2} & {EuroSAT} & {S4A-rand} & {S4A-tiles} & {So2Sat-rand} & {So2Sat-block} \\
  \midrule
  60x60 \textrightarrow original   & 56.72 & 82.85 & \textbf{65.21} & 35.62 & 85.63 & 69.09 \\
  60x60 \textrightarrow 120x120    & 56.72 & \textbf{84.13} & \textbf{65.21} & 35.62 & 94.14 & \textbf{73.71} \\
  60x60 \textrightarrow 264x264    & \textbf{56.95} & 83.97 & 64.83 & \textbf{35.94} & \textbf{94.51} & \textbf{73.84} \\
  \bottomrule
  \end{tabular}
  \label{tab:resize_downstream_60}
\end{table*}

% \begin{table*}[hbt!]
%   \centering
%   \small % Reduce font size to fit within column width
%   \caption{Performance (in \%) of different resizing strategies for downstream datasets evaluated by the \gls{k-NN} protocol when pre-training on SSL4EO with fixed image size. The first image size (left of the arrow) is the centre cropped size of the pre-training dataset, and the second image size (right of the arrow) is the resized downstream image size.}
%   \setlength{\tabcolsep}{7pt} % Adjust padding between columns
%   \begin{tabular}{@{} l *{7}{S[table-format=2.2]} @{}}
%   \toprule
%   Image Size & {BEN-V2} & {EuroSAT} & {S4A-rand} & {S4A-tiles} & {So2Sat-rand} & {So2Sat-cult} \\
%   \midrule
%   60x60 \textrightarrow original   & 56.72 & 82.85 & \bfseries 65.21 & 35.62 & 85.63 & 30.83 \\
%   60x60 \textrightarrow 120x120    & 56.72 & \bfseries 84.13 & \bfseries 65.21 & 35.62 & 94.14 & \bfseries 33.80 \\
%   60x60 \textrightarrow 264x264    & \bfseries 56.95 & 83.97 & 64.83 & \bfseries 35.94 & \bfseries 94.51 & 33.23 \\
%   \bottomrule
%   \end{tabular}
%   \label{tab:resize_downstream_60}
% \end{table*}

\begin{table*}[hbt!]
  \centering
  \small % Reduce font size to fit within column width
  \caption{Performance (in \%) of different resizing strategies for downstream datasets evaluated by the \gls{k-NN} protocol when pre-training on SSL4EO with fixed image size. The first image size (left of the arrow) is the center cropped size of the pre-training dataset, and the second image size (right of the arrow) is the resized downstream image size.}
    \vspace{0.2cm}
  \setlength{\tabcolsep}{7pt} % Adjust padding between columns
  \begin{tabular}{@{} l *{7}{S[table-format=2.2]} @{}}
  \toprule
  Image Size & {BEN-V2} & {EuroSAT} & {S4A-rand} & {S4A-tiles} & {So2Sat-rand} & {So2Sat-block} \\
  \midrule
  120x120 \textrightarrow original   & 58.34 & 82.22 & 65.39 & \textbf{36.21} & 78.76 & 64.80 \\
  120x120 \textrightarrow 120x120    & 58.34 & 85.57 & 65.39 & \textbf{36.21} & 95.06 & 74.57 \\
  120x120 \textrightarrow 264x264    & \textbf{59.18} & \textbf{86.32} & \textbf{66.21} & \textbf{36.27} & \textbf{96.02} & \textbf{75.10} \\
  \bottomrule
  \end{tabular}
  \label{tab:resize_downstream_120}
\end{table*}

% \begin{table*}[hbt!]
%   \centering
%   \small % Reduce font size to fit within column width
%   \caption{Performance (in \%) of different resizing strategies for downstream datasets evaluated by the \gls{k-NN} protocol when pre-training on SSL4EO with fixed image size. The first image size (left of the arrow) is the center cropped size of the pre-training dataset, and the second image size (right of the arrow) is the resized downstream image size.}
%     \vspace{0.2cm}
%   \setlength{\tabcolsep}{7pt} % Adjust padding between columns
%   \begin{tabular}{@{} l *{7}{S[table-format=2.2]} @{}}
%   \toprule
%   Image Size & {BEN-V2} & {EuroSAT} & {S4A-rand} & {S4A-tiles} & {So2Sat-rand} & {So2Sat-cult} \\
%   \midrule
%   120x120 \textrightarrow original   & 58.34 & 82.22 & 65.39 & 36.21 & 78.76 & 30.89 \\
%   120x120 \textrightarrow 120x120    & 58.34 & 85.57 & 65.39 & 36.21 & 95.06 & 34.90 \\
%   120x120 \textrightarrow 264x264    & \bfseries 59.18 & \bfseries 86.32 & \bfseries 66.21 & \bfseries 36.27 & \bfseries 96.02 & \bfseries 35.03 \\
%   \bottomrule
%   \end{tabular}
%   \label{tab:resize_downstream_120}
% \end{table*}

\begin{table*}[hbt!]
  \centering
  \small % Reduce font size to fit within column width
  \caption{Performance (in \%) of different resizing strategies for downstream datasets evaluated by the \gls{k-NN} protocol when pre-training on SSL4EO with fixed image size. The first image size (left of the arrow) is the center cropped size of the pre-training dataset, and the second image size (right of the arrow) is the resized downstream image size.}
  \setlength{\tabcolsep}{7pt} % Adjust padding between columns
  \begin{tabular}{@{} l *{7}{S[table-format=2.2]} @{}}
  \toprule
  Image Size & {BEN-V2} & {EuroSAT} & {S4A-rand} & {S4A-tiles} & {So2Sat-rand} & {So2Sat-block} \\
  \midrule
  264x264 \textrightarrow original   & 57.44 & 80.41 & 64.65 & \textbf{36.45} & 70.68 & 59.15 \\
  264x264 \textrightarrow 120x120    & 57.44 & 84.44 & 64.65 & \textbf{36.45} & 92.87 & 73.45 \\
  264x264 \textrightarrow 264x264    & \textbf{59.08} & \textbf{86.22} & \textbf{66.06} & \textbf{36.61} & \textbf{95.64} & \textbf{75.04} \\
  \bottomrule
  \end{tabular}
  \label{tab:resize_downstream_264}
\end{table*}

% \begin{table*}[hbt!]
%   \centering
%   \small % Reduce font size to fit within column width
%   \caption{Performance (in \%) of different resizing strategies for downstream datasets evaluated by the \gls{k-NN} protocol when pre-training on SSL4EO with fixed image size. The first image size (left of the arrow) is the centre cropped size of the pre-training dataset, and the second image size (right of the arrow) is the resized downstream image size.}
%   \setlength{\tabcolsep}{7pt} % Adjust padding between columns
%   \begin{tabular}{@{} l *{7}{S[table-format=2.2]} @{}}
%   \toprule
%   Image Size & {BEN-V2} & {EuroSAT} & {S4A-rand} & {S4A-tiles} & {So2Sat-rand} & {So2Sat-block} \\
%   \midrule
%   264x264 \textrightarrow original   & 57.44 & 80.41 & 64.65 & \bfseries 36.45 & 70.68 & 28.77 \\
%   264x264 \textrightarrow 120x120    & 57.44 & 84.44 & 64.65 & \bfseries 36.45 & 92.87 & 34.94 \\
%   264x264 \textrightarrow 264x264    & \textbf{59.08 & \textbf{86.22 & \textbf{66.06 & \bfseries 36.61 & \textbf{95.64 & \textbf{35.40 \\
%   \bottomrule
%   \end{tabular}
%   \label{tab:resize_downstream_264}
% \end{table*}

% To split the supplementary pages from the main paper, you can use \href{https://support.apple.com/en-ca/guide/preview/prvw11793/mac#:~:text=Delete%20a%20page%20from%20a,or%20choose%20Edit%20%3E%20Delete).}{Preview (on macOS)}, \href{https://www.adobe.com/acrobat/how-to/delete-pages-from-pdf.html#:~:text=Choose%20%E2%80%9CTools%E2%80%9D%20%3E%20%E2%80%9COrganize,or%20pages%20from%20the%20file.}{Adobe Acrobat} (on all OSs), as well as \href{https://superuser.com/questions/517986/is-it-possible-to-delete-some-pages-of-a-pdf-document}{command line tools}.

\end{document}